\documentclass[onecolumn]{article}
\usepackage{xcolor}
\usepackage{booktabs}
\usepackage{tabularx}
\usepackage{siunitx}

\usepackage{amsmath}
\usepackage{siunitx}
\usepackage{enumitem}
\usepackage[utf8]{inputenc}
\usepackage{tikz}
\tikzstyle{mybox} = [draw=black, very thick, rectangle, rounded corners, inner ysep=5pt, inner xsep=5pt]
\usetikzlibrary{positioning, fit, calc, arrows.meta}
\usepackage{tikz}
\usetikzlibrary{fit, backgrounds}

\usepackage{amsmath, amssymb}
\usetikzlibrary{fit}

\usepackage{algorithmicx}
\usetikzlibrary{calc}
\usepackage{authblk}

\usetikzlibrary{matrix}

\newtheorem{theorem}{Theorem}

\newtheorem{proposition}[theorem]{Proposition}

\begin{document}
\title{Learning stochasticity: a nonparametric framework for intrinsic noise estimation}

\author[1]{G. Pillonetto}
\author[2]{A. Giaretta}
\author[3]{M. Bisiacco}
\affil[1]{Department of Information Engineering, University of Padova, Padova (Italy)}
\affil[2]{Department of Pathology, Cambridge University, Cambridge (UK)}
\affil[3]{Department of Information Engineering, University of Padova, Padova (Italy)}
\date{}                     
\setcounter{Maxaffil}{0}
\renewcommand\Affilfont{\itshape\small}

\maketitle

\begin{abstract}
Understanding the principles that govern dynamical systems is a central challenge across many scientific domains, including biology and ecology. 
Incomplete knowledge of nonlinear interactions and stochastic effects often renders bottom-up modeling approaches ineffective, motivating the development of methods that can discover governing equations directly from data. 
In such contexts, parametric models often struggle without strong prior knowledge, especially when estimating intrinsic noise.
Nonetheless, incorporating stochastic effects is often essential for understanding the dynamic behavior of complex systems such as gene regulatory networks and signaling pathways.
To address these challenges, we introduce Trine (Three-phase Regression for INtrinsic noisE), a nonparametric, kernel-based framework that infers 
state-dependent intrinsic noise from time-series data. Trine features a three-stage algorithm that combines analytically solvable subproblems with a structured kernel architecture that captures both abrupt noise-driven fluctuations and smooth, state-dependent changes in variance. 
We validate Trine on biological and ecological systems, demonstrating its ability to uncover hidden dynamics without relying on predefined parametric assumptions. 
Across several benchmark problems, Trine achieves performance comparable to that of an oracle. Biologically, this oracle can be viewed as an idealized observer capable of directly tracking the random fluctuations in molecular concentrations or reaction events within a cell. The Trine framework thus opens new avenues for understanding how intrinsic noise affects the behavior of complex systems.
\end{abstract}
{{\bf{Keywords:}} nonlinear dynamic systems $|$  stochastic noise $|$ system identification $|$ machine learning  $|$ system biology
\newpage

\section*{Introduction}

Processes that occur within cells are intrinsically stochastic due to the random nature of biochemical reactions. These intrinsic fluctuations often play an important role in cellular functions such as differentiation and drug resistance \cite{Elowitz2000,Shaffer2017,Ripa1998,dennis2002allee}. Stochasticity in gene expression generates phenotypic diversity among genetically identical cells \cite{Elowitz2002, Swain2002}. This variability plays a functional role in processes such as stem cell differentiation and tissue homeostasis \cite{Chang2008, Clayton2007}, immune activation \cite{Feinerman2008}, and microbial survival strategies \cite{Ackermann2015}.
Despite its ubiquity, intrinsic noise is notoriously difficult to quantify experimentally and computationally.

Stochastic models of biochemical networks often take the form of continuous-time discrete Markov processes and are governed  by the Chemical Master Equation \cite{Gillespie1977}. These models have provided important insights into stochastic effects in gene regulation and cell signaling. However, they require knowledge of reaction pathways and rate constants that are often difficult to attain experimentally. Consequently, data-driven system identification methods \cite{Ljung:99,Soderstrom,SpringerRegBook2022} have been developed to reconstruct governing equations directly from time-series observations. These approaches generally rely on parametric assumptions and  struggle to capture complex nonlinear stochastic dynamics---especially when the underlying statistics depend on the system state---without strong prior knowledge of the model structure.
This challenge is especially pronounced in biological systems, where stochasticity is often a critical component of function rather than a mere nuisance. The combination of ill-posedness, limited data, and the need for nonparametric methods renders model inference difficult---yet all the more important.

The problem is closely related to heteroskedastic noise modeling in machine learning \cite{Nix1994,Kendall2017}. 
In this field common noise inference strategies assume that the log-variance of stochastic effects is an unknown function of the system state. This function is then estimated by parameterizing it with a neural network or by modeling its smooth profile as realizations of a Gaussian process 
\cite{Rasmussen,Scholkopf01b,Goldberg1998}. The latter approach, related to kernel-based methods, typically employs approximate inference techniques to achieve computational tractability based on stochastic simulation techniques and iterative methods \cite{lazaro-gredilla2011variational,Kersting2007,Binois2016,Hong2018WeightedHeteroscedasticGP,Tolvanen2014}. However, standard heteroskedastic models often overlook measurement noise---that is, errors introduced through experimental procedures used to observe the system---which is distinct from intrinsic stochasticity generated by the system’s internal dynamics.  Using inappropriate priors---such as overly flexible or heavy-tailed log-normal distributions---can lead to variance profiles that are highly sensitive to observational noise, as we also illustrate in this work. This necessitates a method that leverages the flexibility of nonparametric inference, while simultaneously ensuring robust and interpretable estimates of intrinsic noise.

Here, we present \textbf{Trine} (Three-phase Regression for INtrinsic noisE), the first framework capable of nonparametrically recovering state-dependent intrinsic noise from biological time-series data. Trine overcomes the core limitations of existing methods through a modular, three-stage regression algorithm based on structured kernel estimators. Each stage uses linear estimators with analytical solutions, favoring parsimonious functional forms in cases where unique solutions do not exist~\cite{Wahba:90,PillonettoPNAS}.  A key innovation is a custom-designed kernel that captures the dependence of noise statistics on the system state, as well as the irregularities and discontinuities characteristic of intrinsic fluctuations in molecular systems.

We demonstrate Trine on several case studies---including ecological population dynamics~\cite{Ricker1954}, gene regulatory networks~\cite{Serrano2000,dennis2002allee}, and spiking neurons modeled by the FitzHugh--Nagumo system~\cite{fitzhugh1961impulses}---showing that it accurately reconstructs state-dependent stochastic structure without requiring parametric assumptions or prior system knowledge.  
Crucially, we show that Trine not only outperforms state-of-the-art learning approaches, but also achieves performance comparable to oracle estimators that have access to the true realizations of intrinsic noise. Such information is fundamentally unavailable in realistic experimental settings.
This highlights the method’s ability to recover latent stochastic mechanisms directly from observational data, effectively bridging the gap between theoretical stochastic modeling and experimental measurability.

\begin{figure}[htp]
\centering
\scalebox{0.9}{
\centering

\def\yshiftTop{1.5cm}

\begin{tikzpicture}[
  node distance=2.8cm and 1cm,
  box/.style = {
    draw, 
    rounded corners, 
    minimum width=10cm, 
    minimum height=2cm, 
    align=center,
    font=\sffamily\small,
    fill=blue!10
  },
  arrow/.style = {-{Latex[length=3mm]}, thick},
  image/.style = {minimum height=1.8cm, minimum width=2.5cm},
  every node/.style={font=\sffamily}
]


\node[box, fill=gray!10, minimum width=4.3cm, minimum height=1.6cm] (equation) at (0,\yshiftTop) {\large
  $\displaystyle 
\dot{x} = f(x) + g(x) w(t) 
  $\\[5pt] 
   \large $\displaystyle
  y(t) = x(t) + e(t)
  $
};

\node[image, right=1cm of equation] (noisydata) {
  \begin{tikzpicture}[x=0.3cm, y=0.8cm]

    \node at (6,-1) {\footnotesize Time};

    \begin{scope}[yshift=-0.6cm]
      \draw[->, thick] (0,0) -- (12,0);
      \draw[->, thick] (0,0) -- (0,1.6);

      \coordinate (P0) at (0.3, {1 + 0.3*sin(0.3 r) + 0.2*(rand)});
      \foreach \i in {1,...,39} {
        \pgfmathsetmacro{\x}{0.3 + 11*(\i)/39}
        \pgfmathsetmacro{\y}{0.5 + 0.3*sin(\x r) + 0.2*(rand)}
        \coordinate (P\i) at (\x,\y);
      }
      \foreach \i in {0,...,39} {
        \fill[blue!70] (P\i) circle[radius=0.04cm];
      }
      \draw[red, thick, smooth] plot [smooth] coordinates {
        (P0) (P1) (P2) (P3) (P4) (P5) (P6) (P7) (P8) (P9)
        (P10) (P11) (P12) (P13) (P14) (P15) (P16) (P17) (P18) (P19)
        (P20) (P21) (P22) (P23) (P24) (P25) (P26) (P27) (P28) (P29)
        (P30) (P31) (P32) (P33) (P34) (P35) (P36) (P37) (P38) (P39)
      };
      \node at (6,1.5) {\footnotesize \textcolor{gray}{Noisy states}};
    \end{scope}
  \end{tikzpicture}
};

\draw[
  -{Latex[length=4mm, width=2.5mm]},
  ultra thick,
  draw=blue!60!black,
  line cap=round
] 
  ($(equation.south east)!0.5!(noisydata.south west) + (0,\yshiftTop + 0.3cm)$) -- 
  ($(equation.south east)!0.5!(noisydata.south west) + (0,\yshiftTop - 1.58cm)$);

\coordinate (step1pos) at ($(equation.south east)!0.5!(noisydata.south west) + (0,\yshiftTop - 2.2cm)$);


\node[box, fill=cyan!20, minimum width=9.5cm, minimum height=2.5cm, anchor=north] (step1) at (step1pos) {
  \begin{minipage}[t]{9cm}
    \centering
    \vspace*{-10pt}
    \textbf{\large Step 1: Intrinsic Noise Signs Estimation} \\
    \vspace{6pt}
    \hspace{-3cm} \textit{Smooth kernel} \\
    \hspace{-3cm} for the deterministic system part $f$
  \end{minipage}
};

\node[image, above=-0.7cm of step1, xshift=3cm, anchor=north] (k1) {
  \begin{tikzpicture}
    \draw[thick, smooth, domain=-1.5:1.5, samples=100] plot(\x,{0.5 + 0.8*exp(-\x*\x)});
    \node at (0,1.5) {\footnotesize \textcolor{gray}{Smooth Kernel}};
  \end{tikzpicture}
};
\node[image, below=-0.4cm of step1, xshift=2.4cm, anchor=north] (s1) {
  \begin{tikzpicture}
    \node at (0.1,0.3) {\textcolor{red}{$+$}\quad\textcolor{blue}{$-$}\quad\textcolor{red}{$+$}\quad\textcolor{red}{$+$}\quad\textcolor{blue}{$-$}};
    \node at (0.1,-0.3) {\footnotesize \textcolor{gray}{Estimates of Intrinsic Noise Signs}};
  \end{tikzpicture}
};

\node[box, fill=orange!20, minimum width=9.5cm, minimum height=2.5cm, below=1.8cm of step1] (step2) {
  \begin{minipage}[t]{9cm}
    \centering
    \vspace*{-10pt}
    \textbf{\large Step 2: Intrinsic Noise Estimation} \\
    \vspace{6pt}
    \hspace{-3cm} \textit{Structured and discontinuous kernel} \\
    \hspace{-3cm} for intrinsic noise realizations
  \end{minipage}
};
\node[image, above=-0.8cm of step2, xshift=3cm, anchor=north] (k2) {
  \begin{tikzpicture}
    \draw[thick, smooth, domain=-1.5:1.5, samples=100] plot(\x,{0.8*exp(-\x*\x)});
    \draw[thick] (0,0.8) -- (0,1.2);
    \node at (0,0.9) {\footnotesize \textcolor{gray}{Discontinuous Kernel}};
  \end{tikzpicture}
};
\node[image, below=0.15cm of step2, xshift=2.5cm, anchor=north] (noise_realizations) {
  \begin{tikzpicture}[x=0.37cm, y=0.6cm]
    \draw[-] (0.5,-0.2) -- (12,-0.24);
    \node at (6.45,-0.01) {\footnotesize \textcolor{gray}{Estimates of Intrinsic Noise}};
    \def\n{30}
    \def\r{0.1}
    \foreach \i in {1,...,\n} {
      \pgfmathsetmacro{\x}{1 + 10.5*(\i-1)/(\n-1)}
      \pgfmathsetmacro{\var}{0.01*\x}
      \pgfmathsetmacro{\y}{rand*2 - 1}
      \pgfmathsetmacro{\yscaled}{\y*sqrt(\var)}
      \fill[blue!70] (\x, \yscaled) circle[radius=\r];
    }
  \end{tikzpicture}
};

\node[box, fill=green!20, minimum width=9.5cm, minimum height=2.5cm, below=1.8cm of step2] (step3) {
  \begin{minipage}[t]{9cm}
    \centering
    \vspace*{-10pt}
    \textbf{\large Step 3: Intrinsic Noise Variance Estimation} \\
    \vspace{6pt}
    \hspace{-3cm} \textit{Smooth kernel} \\
    \hspace{-3cm} for intrinsic noise profile $g$
  \end{minipage}
};
\node[image, above=-0.7cm of step3, xshift=3cm, anchor=north] (k3inside) {
  \begin{tikzpicture}
    \draw[thick, smooth, domain=-1.5:1.5, samples=100] plot(\x,{0.5 + 0.8*exp(-\x*\x)});
    \node at (0,1.5) {\footnotesize \textcolor{gray}{Smooth Kernel}};
  \end{tikzpicture}
};

\node[below=1.85cm of step3] (arrowend3) {}; 
\draw[
  -{Latex[length=4mm, width=2.5mm]},
  ultra thick,
  draw=blue!60!black,
  line cap=round
] 
($(step3.south) + (0,-0.63cm)$) -- (arrowend3);

\node[image, below=-0.9cm of arrowend3, xshift=0.6cm] (k3result) {
  \begin{tikzpicture}[scale=1]
    \draw[->, thick, draw=black!80] (-2.2,0) -- (2.5,0) node[pos=1, below=-12pt] {\footnotesize \textbf{system state}};
    \draw[->, thick, draw=black!80] (-2,0) -- (-2,2);

    \draw[ultra thick, smooth, domain=-2:2, samples=100, draw=blue!60!cyan] 
      plot(\x,{0.3 + 0.7*(1 - exp(-(\x+2)/1.5))});

    \node at (0,2.3) {\footnotesize \textcolor{blue!60!black}{\textbf{Intrinsic Noise Variance}}};
  \end{tikzpicture}
};

\draw[
  -{Latex[length=4mm, width=2.5mm]},
  ultra thick,
  draw=blue!60!black,
  line cap=round
] 
(step1) -- (step2);
\draw[
  -{Latex[length=4mm, width=2.5mm]},
  ultra thick,
  draw=blue!60!black,
  line cap=round
] 
(step2) -- (step3);

\node[
  draw=blue!60!black,
  rounded corners=5pt,
  line width=1pt,
  dashed,
  dash pattern=on 5pt off 2pt,
  inner sep=0.6cm,
  fill=blue!5,
  fill opacity=0.07,
  fit=(step1)(step2)(step3)
] (dashedbox) {};

\node[
  font=\sffamily\bfseries\large, 
  text=black, 
  anchor=north, 
  yshift=-0.05cm
] at (dashedbox.north) [text=red]{Intrinsic Noise Model Estimator};

\end{tikzpicture}
}
\caption{\footnotesize
\textbf{Structured Intrinsic Noise Estimation via Trine}.
The proposed method decomposes the estimation of structured intrinsic noise into three sequential phases using Gaussian Process Regression (GPR) with customized kernels: \emph{Step 1---Sign Estimation:} A GPR model with a smooth kernel is used to describe the deterministic component $f $ of the system dynamics. The residuals from this model are then used to estimate the signs of the intrinsic noise realizations, capturing directional information. \emph{Step 2---Intrinsic Noise Realization:} Based on the estimated signs, a second GPR model is employed, featuring a structured kernel that captures both the discontinuities in the intrinsic noise realizations and the smooth variation of the variance profile with respect to the system state. This model is used to recover the realizations of the intrinsic noise. \emph{Step 3---Noise Variance Profiling:} The absolute residuals from Step 2, appropriately scaled, are modeled using a third GPR with a smooth kernel to estimate the state-dependent noise standard deviation profile. Overall, this modular approach allows for flexible modeling of nonhomogeneous, state-dependent noise in complex dynamical systems by decoupling directionality, structure, and variance.
}
\label{Fig1}
\end{figure}
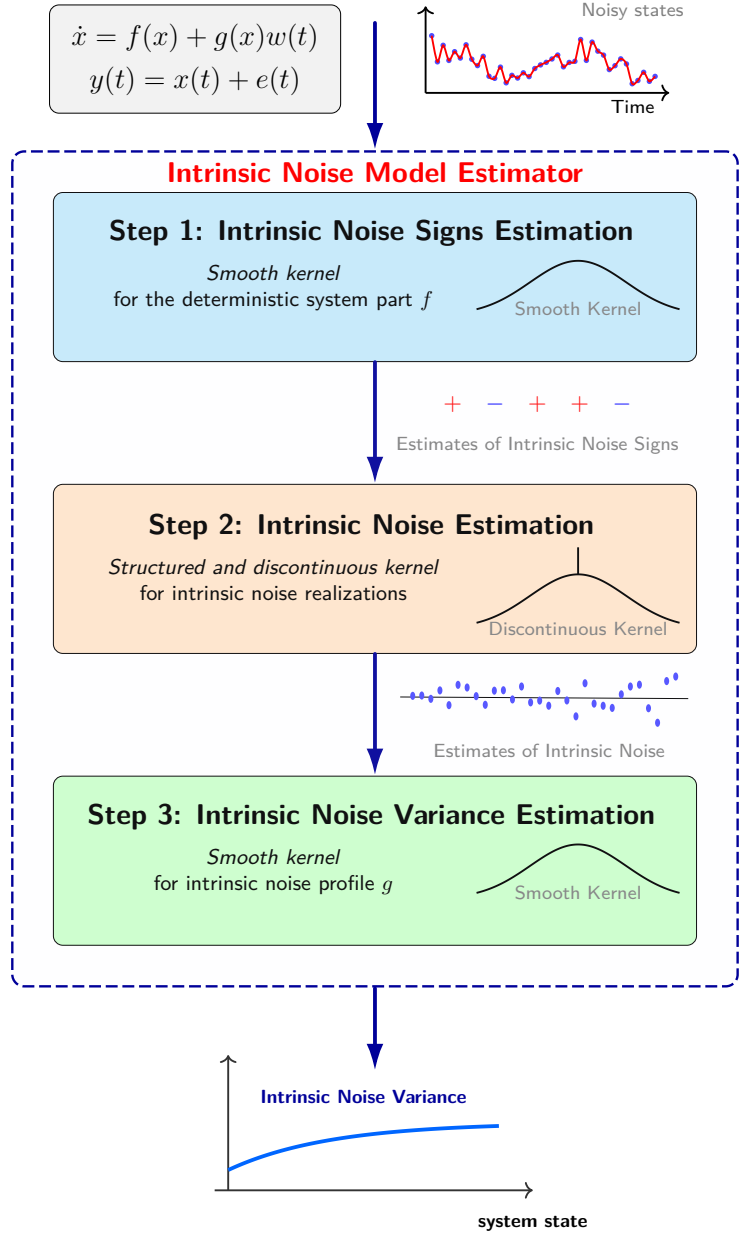

\section*{Intrinsic Noise estimation}

\paragraph{Stochastic state-space models.}
We consider stochastic dynamical systems subject to both intrinsic and experimental noise. In the continuous-time setting, the system is described by the following state-space model:
\begin{align}
\dot{x}(t) &= f(x(t)) + g(x(t)) \cdot w(t), \label{eq:state-eq} \\
y_k &= x(t_k) + e_k, \qquad k = 1, \dots, N, \label{eq:output-eq}
\end{align}
where $x(t) \in \mathbb{R}^n$ denotes the system state at time $t$, and $y_k \in \mathbb{R}^n $ are noisy observations collected at discrete time points $t_k $. The function $f$ describes the deterministic drift, while the noise term $g \cdot w$ represents the intrinsic noise, reflecting internal stochasticity dependent on the system state. 
This term includes a matrix-valued function $g(x)$ and a random vector $w(t)$, whose components are independent Gaussian white noise processes. A diagonal $g(x)$ corresponds to the case where the noise inputs driving the components of $x(t)$ are uncorrelated.
Finally, the term $e_k$ accounts for measurement noise. This formulation thus captures the two primary sources of uncertainty present in many biological and ecological systems. The first is intrinsic noise, arising from internal stochastic mechanisms such as gene expression bursts or random molecular interactions within a cell; the second is measurement noise, caused by imperfections in the observation process, such as sensor inaccuracies.

The two central functions to be estimated from a finite set of noisy data are the deterministic drift $f$ and the noise strength $g$. This estimation task is challenging due to the nonlinear structure of these functions. Additionally, $g$ appears within a stochastic term in the dynamics, further complicating the problem. As a result, the problem is fundamentally ill-posed: infinitely many combinations of $f$ and $g$ can explain the observed data equally well, particularly in noisy settings. A common strategy for regularization is to adopt a parametric formulation, reducing the problem to estimating a finite number of parameters within predefined functional families. While this improves tractability, it imposes strong prior assumptions on the structure of $f$ and $g$, which are often difficult to justify in settings with poorly understood or highly nonlinear mechanisms.

\paragraph{The TRINE algorithm} 
To estimate the components of this system from data, we adopt a nonparametric approach based on kernel methods---a flexible and principled tool that regularizes the estimation problem without imposing explicit functional forms \cite{Scholkopf01b}. These methods encode prior assumptions via a symmetric positive semidefinite function $\mathcal{K} \colon \mathbb{R}^n \times \mathbb{R}^n \rightarrow \mathbb{R}$, where 
$\mathcal{K}(x_a,x_b)$ quantifies the similarity between states $x_a$ and $x_b$. This means that pairs of states that are ``close'' according to some metric tend to have similar values of the unknown function, i.e., $f(x_a) \approx f(x_b)$.
For example, if $\mathcal{K}$ is continuous, any function drawn from the associated model will also be continuous. Defining a kernel implicitly determines a potentially infinite set of basis functions---specifically, the kernel sections---that span the function space used for estimation. This construction not only provides great flexibility in approximating complex functions but also induces an implicit regularization on the expansion coefficients. This can be interpreted probabilistically: the kernel defines the covariance of a Gaussian process prior over functions, favoring those with higher prior probability (e.g., smoother if the kernel is regular) \cite{Rasmussen}. This naturally promotes simpler functional forms when multiple solutions fit the data equally well, thus helping to avoid overfitting without explicitly restricting the model class.

A widely used kernel to encode smoothness is the \emph{Gaussian kernel}, defined as
$$
\mathcal{K}(x_a,x_b) = \lambda \exp\left(-\frac{\|x_a - x_b\|^2}{2\ell} \right),
$$
where $\lambda > 0$ and $\ell > 0$ are hyperparameters controlling the amplitude and smoothness of the estimate. These parameters are typically learned from data. Gaussian kernels are universal, in the sense that they can approximate any continuous function arbitrarily well \cite{Micchelli2006}, and are particularly effective when the target function is expected to be smooth. 
In the Trine method, the Gaussian kernel is employed directly in some stages and also serves as a foundation for designing custom kernels in others, as described below.\\

Trine is a three-stage estimator designed to reconstruct not only the drift $f$ and the standard deviation profile $g$, but also the intrinsic noise realizations $g \cdot w$,  which are recovered in the middle stage of the algorithm. Crucially, this intermediate quantity---the intrinsic noise---is not merely a byproduct of the model: it is a central component of the algorithm. From a modeling perspective, it provides direct insight into the stochastic mechanisms driving the system, which is key in domains where noise itself plays a functional role. From a computational perspective, estimating $g \cdot w$ is instrumental, as it enables the subsequent inference of the state-dependent variance defined by $g$.

The algorithm is graphically depicted in Fig.~\ref{Fig1}, which illustrates how this three-phase framework enables the separate and interpretable modeling of the structure and scale of the intrinsic noise. The modular steps of Trine can be briefly described as follows. 
First, the drift function $f$ is estimated using Gaussian Process Regression (GPR) with a Gaussian kernel (see the upper blue box and smooth kernel curve in Fig.~\ref{Fig1}). A simplified intrinsic noise model with constant variance is assumed at this stage, and subsequently refined in the following steps. From the residuals of this step, the signs of the intrinsic noise realizations are inferred, capturing their directional component (see the red/blue sign estimates below the upper blue box in Fig.~\ref{Fig1}). Second, a specialized GPR model with a newly introduced structured, discontinuous kernel reconstructs the intrinsic noise realizations (see the middle orange box and discontinuous kernel curve in Fig.~\ref{Fig1}). This step leverages the sign estimates while preserving the smooth structure of the variance. Third, the absolute residuals from the previous step are appropriately scaled and used as input to a final GPR with a Gaussian kernel to estimate the state-dependent intrinsic noise standard deviation (SD) (see the bottom green box and smooth kernel curve above the final noise variance profile in Fig.~\ref{Fig1}).  This last step can be further adapted to enforce structural constraints such as monotonicity or concavity, embedding biologically motivated priors on the variance profile.

The pseudocode of Trine is provided in Appendix, together with the theoretical motivations underlying its structure---including the derivation of the novel kernel used in the second phase.

\section*{Results}

\subsection*{Population dynamics: Ricker Model with Allee Effect}

The Ricker model is a classic discrete-time population model 
developed for fish stock dynamics \cite{Ricker1954}. Like the logistic map, it describes density-dependent growth, but with an exponentially decreasing per-capita rate, making it suitable for species with strong overcompensation, where populations drop sharply after exceeding carrying capacity. Its rich nonlinear behavior has made it a standard tool for studying chaos, bifurcations, and noise effects in ecological systems \cite{May1976}. Recent developments include stochastic competition models \cite{Dallas2021}, stock--recruitment models with environmental covariates \cite{GaoWang2022}, and theoretical analyses \cite{Elaydi2024}.

In our example, the deterministic drift also incorporates an Allee effect:
$$
f(x_k) = x_k^2 \, e^{r(1 - x_k)},
$$
which introduces a positive relationship between population size and growth at low densities. Populations below a critical threshold then face negative growth, increasing extinction risk \cite{dennis2002allee}. Incorporating the Allee effect enables the study of bistability between extinction and survival, especially under stochastic influences.

As discussed in \cite{dennis2002allee}, in stochastic population models the type of random fluctuations determines the appropriate diffusion approximation: demographic noise may produce variance proportional to the population size $x$, whereas environmental noise leads to variance proportional to $x^2$. We tested the Trine approach under both noise types, successfully recovering the system dynamics in each case. Here, we present results for the system
$$
x_{k+1}=f(x_k)+g(x_k)w_k
$$
with $f$ given above and demographic noise characterized by 
$$
g(x) = \sqrt{0.3^2 + 0.05^2 \, x},
$$
as in \cite{Bashkirtseva2014}.

Simulations were performed with $r = 2.5$. 
In this and all subsequent examples, the standard deviation of the output noise $e_k$ is set to a fixed fraction of the true state $x_k$, such that the norm of the output noise realizations is approximately 30--40\% of the norm of the intrinsic noise. The top panel of Fig.~\ref{FigRicker} presents a simulated trajectory whose dynamics display irregular oscillations. The bottom-left panel shows that the Trine estimate of $g(x)$ (blue) closely matches the true profile (red), while the drift estimate (bottom-right panel) is also accurate, with only slight degradation near the upper boundary of the state space---an area less frequently visited by the system.

\begin{figure*}[h]
	\begin{center}
        { \includegraphics[scale=0.7]{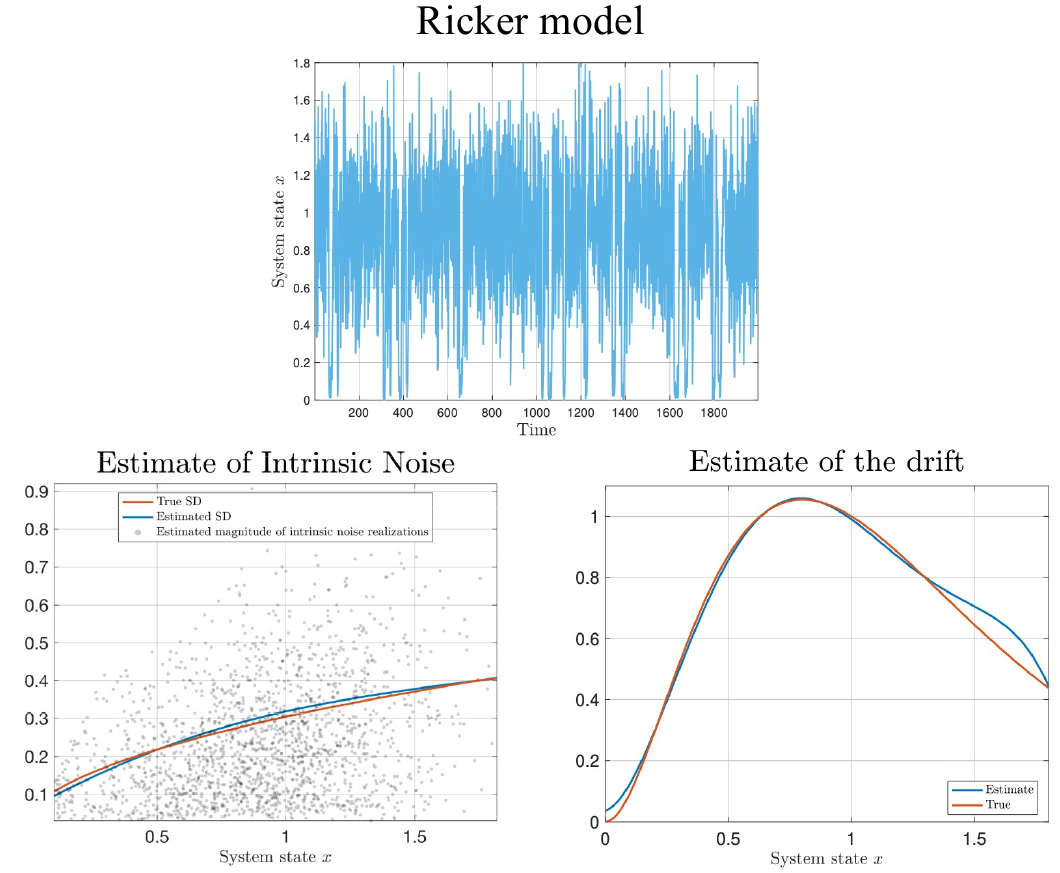}} 
		\caption{Results for Ricker model. The top panel show the simulations of the system using $r=2.5$ and intrinsic noise SD
		$g(x)=\sqrt{0.3^2+0.05^2x}$ \cite{Bashkirtseva2014}. The bottom left panel shows the estimated absolute values of the intrinsic noise realizations returned by the second stage of Trine (grey dots), the true $g$ function (red) and the estimated one returned by Trine (blue). The bottom right panel displays the true deterministic part $f$ (red) and the estimate by Trine (blue).}  \label{FigRicker}
	\end{center}
\end{figure*}

\subsection*{FitzHugh--Nagumo Model with Stochastic Noise}

The FitzHugh--Nagumo (FHN) model is a paradigmatic reduction of the Hodgkin--Huxley equations \cite{hodgkin1952}, designed to capture the essential dynamical features of excitable systems such as neurons and cardiac cells. By simplifying the high-dimensional biophysical description into a two-dimensional system, the FHN model provides a tractable framework for studying excitability, oscillations, and wave propagation in both single cells and extended media \cite{fitzhugh1961impulses,nagumo1962active}. Its minimal formulation, consisting of a fast activator variable $V(t)$ and a slower recovery variable $W(t)$, has made it one of the most widely used models for theoretical and computational investigations of excitable dynamics. In a neuronal context, $V$ corresponds to the membrane potential, which exhibits rapid, nonlinear excursions during action potentials, while $W$ accounts for slower recovery mechanisms, such as sodium channel inactivation and potassium activation, that return the system to rest.

To capture the intrinsic variability present in real biological systems—arising from ion channel stochasticity, synaptic inputs, or environmental fluctuations—a stochastic extension of the FHN model is considered, in which both variables are subject to multiplicative, state-dependent noise \cite{lindner2004effects,schmid2001stochastic}:
\begin{align}
\dot V &= V - \frac{V^3}{3} - W + I_{\rm ext} + \sigma_V |V|^\alpha \eta_V(t), \\
\dot W &= \epsilon (V + a - b W) + \sigma_W |W|^\beta \eta_W(t),
\end{align}
where $\eta_V(t)$ and $\eta_W(t)$ are independent Gaussian white noise processes. The deterministic dynamics are governed by the parameters $\epsilon$, $a$, $b$, and $I_{\rm ext}$, which control timescale separation, excitability, and external input. The stochastic components are characterized by $\sigma_V$, $\sigma_W$, $\alpha$, and $\beta$, determining the noise intensity and its state dependence. This stochastic formulation enables the investigation of phenomena such as noise-induced oscillations, coherence resonance, and threshold modulation \cite{tuckwell2005introduction,Yamakou2019,Zhang2022,Bao2024}.

Simulations of the stochastic FHN system were performed using the Euler--Maruyama method with a time step of $0.01$. The system was placed in an excitable regime 
and a total of 2000 data points for $V$ and $W$ were sampled at intervals of $\Delta t = 0.1$,
see the top panel of Fig.~\ref{FigFHN}. To emulate realistic experimental conditions, measurement noise was added to both the system states. As in the previous case studies its norm turns out to be approximately one third of that of the intrinsic process noise.
The Trine algorithm was used to estimate the stochastic structure from this noisy time series. We set $\sigma_V = 0.1$ and $\sigma_W = 0.05$. Then,
to assess its robustness, the method was applied across a range of values for the noise exponents $\alpha$ and $\beta$, consistently achieving high performance. 
As a representative example, we focus here on the non-polynomial case $\alpha = \beta = 0.8$. Although this form of state-dependent noise lies outside the class of polynomial functions, Trine successfully recovers it in a fully nonparametric fashion. This is illustrated in Fig.~\ref{FigFHN}: the bottom-left panel shows the true state-dependent standard deviation driving $\dot V$, while the bottom-right panel displays the estimate inferred by Trine. The close match between them 
confirms the method’s ability to accurately identify complex, nonlinear noise structures directly from data.
Moreover, the nonparametric estimate offers practical advantages for experimentalists. From the estimated standard deviation function, a biologist can readily infer that the stochastic forcing in $\dot V$ depends primarily on the state variable $V$, with negligible or no dependence on $W$. In addition, the strong symmetry of the recovered profile suggests that the noise intensity is governed mainly by $|V|$, supporting a hypothesis of modulus-based dependence. Such qualitative insights, directly extracted from data without imposing a predefined functional form, can guide the formulation of more refined mechanistic models or inspire targeted experimental investigations.

\begin{figure*}[h]
	\begin{center}
\includegraphics[scale=0.32]{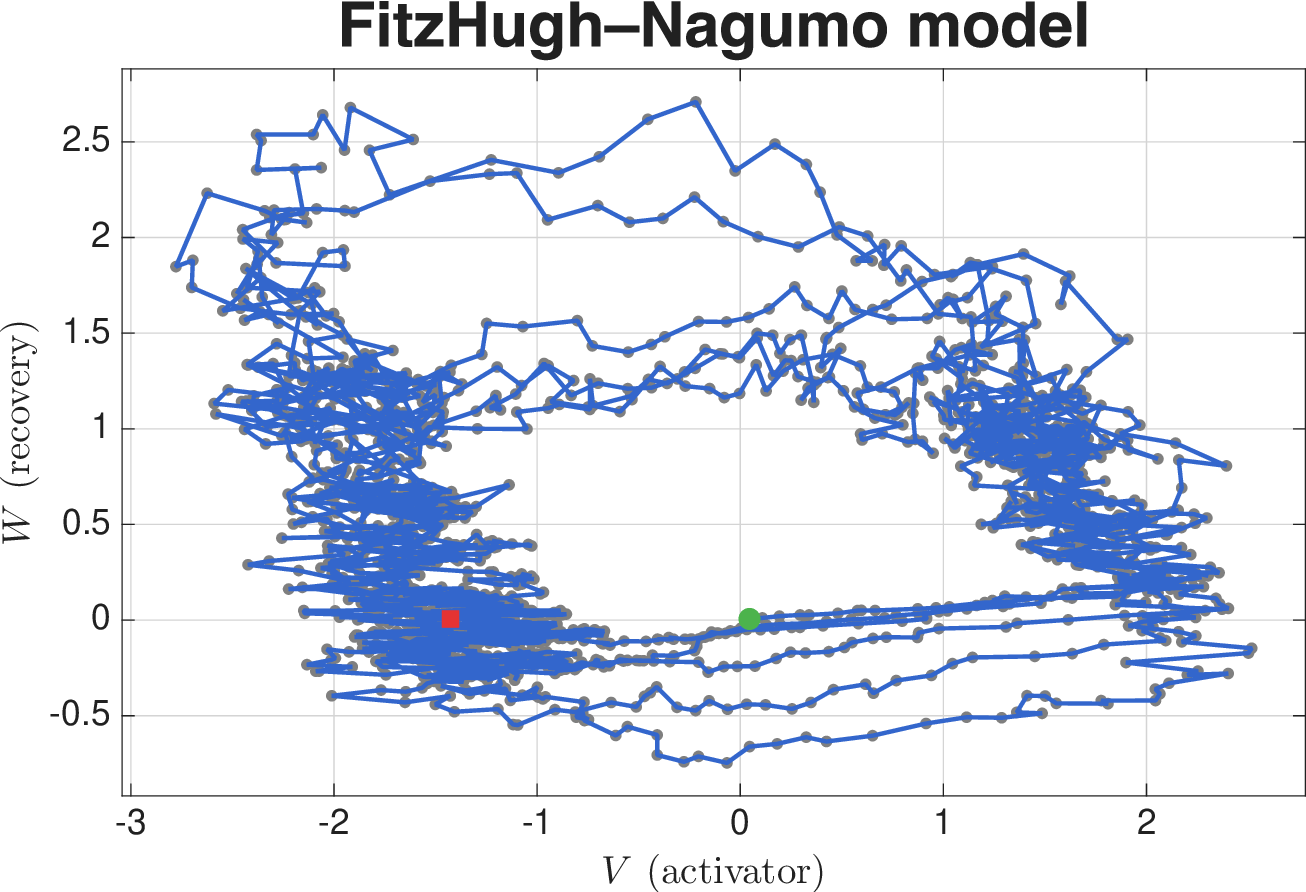}\\
			\includegraphics[scale=0.44]{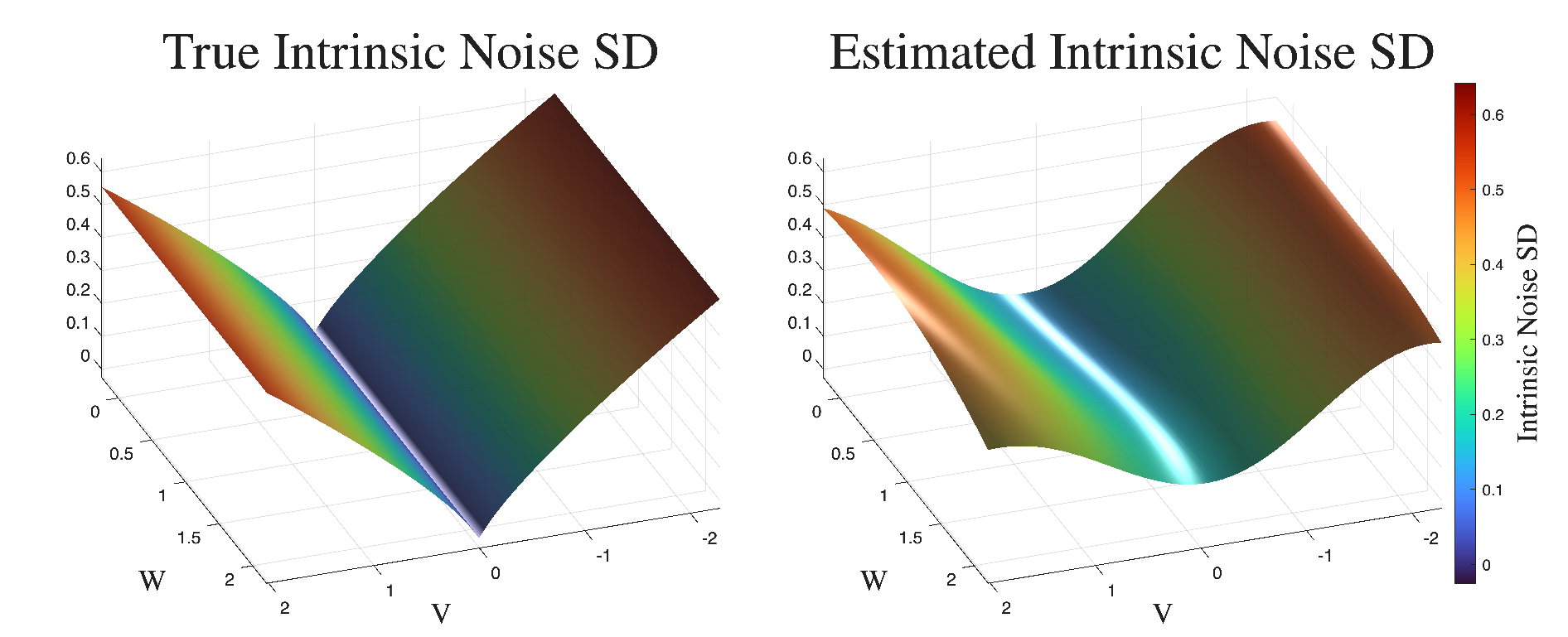}
	\end{center}
    \caption{Results for FitzHugh--Nagumo  model. The top panel shows a representative simulated trajectory in the 2D plane $(V,W)$
     obtained using parameters $\epsilon = 0.08$, $a = 0.7$, $b = 0.8$, $I_{\rm ext} = 0.5$, $\sigma_V = 0.1$, $\sigma_W = 0.05$ and $\alpha = \beta = 0.8$. Stochastic dynamics qualitatively show two basins of attraction. The left-bottom panel shows the true state-dependent standard deviation driving $\dot{V}$, while the right-bottom panel displays its Trine estimate. 
    }  \label{FigFHN}
\end{figure*}

\subsection*{Gene regulatory networks}

\paragraph{Self promoter}
The self-promoter is a canonical stochastic gene regulatory model in systems biology, where a gene activates its own transcription through positive feedback. This simple architecture can generate bistability, hysteresis, and noise-induced switching, providing a minimal framework to study how intrinsic noise shapes phenotypic variability \cite{Kaern2005,Elston2001,Groisman2008}. Positive autoregulation is widespread: it drives the lysis–lysogeny decision in bacteriophage $\lambda$ \cite{Arkin1997}, has been engineered in synthetic circuits to demonstrate robust bistable switching in both E. coli \cite{Serrano2000} and yeast \cite{Serrano2001}, and is a recurrent strategy in virology. For instance, the HIV Tat loop produces bimodal dynamics that control latency versus replication \cite{Weinberger2005}, while HPV early genes sustain their own promoter activity to stabilize early genes programs \cite{Giaretta2020,Giaretta2019}. Beyond viruses, self-promoting loops also appear in mammalian regulation, e.g. in $c$-Myc and $p53$ feedback, reinforcing proliferation or stabilizing cell-fate decisions. Thus, the self-promoter model provides a minimal yet generalizable framework to explore how feedback and noise jointly govern cellular decision-making.

We consider a self-promoter system under a fast promoter fluctuation regimen, which leads to a non-switching SDE formulation \cite{Elston2001}. 
The drift term 
\begin{equation}\label{Eq1}
f(x) = \frac{b a_0 + x^2}{b + x^2} - x - \frac{2xb (a_0-1) \big[ ((a_0-2)+x)x^2+b(x-a_0) \big]}{\kappa (b+x^2)^4}.
\end{equation}
describes the deterministic balance of the system. Here, $x$ denotes the mRNA or protein concentration, $a_0$ controls the basal expression level, $b$ represents the binding affinity of the activator to the promoter, and $\kappa$ is the promoter switching rate. The diffusion term 
\begin{equation}\label{Eq2}
g(x) = \sqrt{\frac{1}{m_0} \left(\frac{b(a_0+x)+x^2(1+x)}{b+x^2}\right) + \frac{bx^2 (a_0-1)^2}{\kappa (b+x^2)^3}}.
\end{equation}
quantifies intrinsic noise. Its amplitude is controlled by $m_0$, which scales fluctuations from synthesis and degradation, and by $\kappa$, which governs noise from stochastic promoter switching, with smaller values indicating stronger noise. As a result, noise is suppressed at low and high expression levels but peaks at intermediate concentrations, where promoter on/off transitions occur most frequently.

The self-promoter model described by \eqref{Eq1} and \eqref{Eq2} thus depends on four key parameters with clear biological interpretation \cite{Elston2001}. The scalar $a_0$ defines the basal activity level, $b$ the strength of positive feedback, $m_0$ the typical steady-state protein copy number, which sets the scale of the dynamics. Finally, $\kappa$ quantifies stochastic fluctuations.  Different parameter sets yield distinct dynamical regimens, ranging from monostability to bistability or noise-dominated behavior. Parameters used in the simulations are reported in Fig. \ref{FigGene1D}.

Identification data consist of 
1000 points collected from four independent experiments. The stochastic differential equation is simulated using the Euler--Maruyama method with a time step of $0.01~\text{min}$.  
Data are sampled at intervals $\Delta_t = 0.01$, though other values of $\Delta_t$ within the range $[0.01, 0.1]$ have also been tested, yielding results consistent with those presented below.  
The rationale behind the choice of measurement noise statistics is the same as in the previous example, ensuring that the output noise norm is approximately 30--40\% of the intrinsic noise norm.
The top panel of Fig.~\ref{FigGene1D} shows a representative simulated trajectory, characterized by pronounced burst-like stochastic behavior.  
The bottom panel illustrates how Trine's estimate of $g(x)$ (blue) closely matches the true diffusion profile (red). As expected, the profile is low at very small and very large protein concentrations, and reaches a maximum at intermediate levels---precisely where promoter switching between active and inactive states occurs most frequently.
Most data points are concentrated near zero expression levels, while intermediate states are relatively underrepresented.  
Nevertheless, the kernel-based estimation accurately reconstructs the non-monotonic shape of the diffusion profile.
Even without assuming a parametric model, Trine thus provides an intuitive signature of stochastic promoter dynamics, offering modelers a direct handle on how intrinsic noise shapes cell-to-cell variability.

\paragraph{Mutual repressor: Toggle switch}
The mutual repressor, or genetic toggle switch, is a minimal circuit where two genes inhibit each other, generating bistability and mutually exclusive expression states. Its first synthetic implementation in E. coli \cite{Gardner2000} was a milestone in synthetic biology, later formalized in the stochastic setting to show how intrinsic noise drives switching \cite{Elston2001}. Since then, the toggle switch has become a paradigmatic systems biology motif for studying stochastic switching, hysteresis, and nonlinear regulation \cite{Cherry2000,Kaern2005}. Natural analogues include the PU.1–GATA1 antagonism in hematopoietic lineage choice \cite{Peterson2009} and the miR-200/ZEB feedback in EMT \cite{Lu2013}. Recent studies have extended this picture: synthetic toggles have been used to probe spatiotemporal dynamics in E. coli \cite{Barbier2020}, while Jagged–Delta asymmetry in Notch signaling generates hybrid sender/receiver states through toggle-like dynamics \cite{Boareto2015}. More broadly, mutual repression motifs are increasingly viewed as modular building blocks for complex architectures, capable of producing oscillations, multistability, and spatial patterning \cite{Guantes2008,Balazsi2011}. This modular perspective emphasizes the central role of the toggle switch as both a theoretical archetype and a practical design principle across natural and engineered systems.

In our example, we model the toggle switch using a two-dimensional stochastic differential equation (SDE), written in the compact vector form given in \eqref{eq:state-eq},  
where $x \in \mathbb{R}^2$ contains the normalized concentrations of two proteins.  
The diffusion matrix $g(x) \in \mathbb{R}^{2 \times 2}$ captures stochastic dynamics through two independent intrinsic noise sources, each associated with the expression of one of the mutually repressive proteins. The state-dependent standard deviations, represented by its diagonal entries, quantify fluctuations specific to each protein.
Importantly, $g(x)$ has a highly intricate parametric structure, reflecting complex, state-dependent interactions that make its analytic form nontrivial, as can be seen from the full modeling details provided in Appendix.

The simulation setup, including the measurement noise characteristics, follows that of the previous case study.  
We focus here on estimating the standard deviation of the intrinsic noise component affecting the dynamics of the first protein, using 1000 noisy measurements of both state variables.  
The results are representative of those obtained from extensive Monte Carlo simulations with independent noise realizations and varying sampling intervals $\Delta_t$.
The top panel of Fig.~\ref{FigGene2D} displays a representative trajectory of the two protein concentrations in the 2D state space, clearly revealing two basins of attraction---a hallmark of bistable systems.  
The bottom panels show, respectively, the true (left) and estimated (right) magnitudes of the intrinsic noise and their corresponding standard deviation profiles, computed using $\Delta_t = 0.01$.  
The left panel shows the ground truth noise realizations and state-dependent SD, while the right panel demonstrates that the Trine estimate closely matches the true profile. Despite the simulated trajectory not uniformly exploring the state space---most samples are concentrated in the two stable basins, with fewer in the transition region---Trine still reconstructs the noise SD surface accurately. The nonparametric estimate of the diffusion offers a direct view of how fluctuations vary across the space, highlighting regions that are most sensitive to stochastic fluctuations in state switching and cell-fate decisions.

\subsection*{Comparison with other approaches} 

To benchmark Trine, we compared it with three alternative estimators using extensive Monte Carlo simulations across four systems presented earlier.

The first comparator is an {\bf Oracle} estimator. In statistics, an “oracle” often refers to a method that has access to privileged information unavailable in practice, providing a theoretical performance upper bound. Here, we define the oracle as having access to the true realizations of intrinsic noise. Biologically, this corresponds to an idealized observer capable of directly tracking random fluctuations in molecular concentrations or reaction events—information hidden in real experiments. The oracle thus obtains noiseless measurements of $g(x)w(t)$ in \eqref{eq:state-eq} and estimates the noise profile using Trine’s third stage, providing a strong upper bound on the performance achievable by any data-driven method.

The second comparator is a classical literature method based on a log-normal Gaussian process prior, commonly used to model state-dependent variance~\cite{Goldberg1998}. Specifically, we consider \emph{Most Likely Heteroscedastic Gaussian Process regression} (\textbf{MLH-GP})~\cite{Kersting2007}, which iteratively estimates the mean function (deterministic system part) and the state-dependent variance profile. This serves as a well-established baseline for nonparametric modeling of state-dependent noise, as demonstrated by its continued study and applications in recent years \cite{Boukouvalas2014,Binois2019,Chung2019,Faza2025}. However, like other heteroscedastic noise methods, it does not explicitly account for observational noise. We therefore implemented an extended version incorporating known measurement noise variances during mean and variance estimation. 

The third comparator is denoted by  \textbf{Trine$^{\mathbf{u}}$}, a Trine variant where the third stage---estimating the intrinsic noise variance---relies directly on intrinsic noise estimates from the GP model in the first stage, bypassing the structured kernel of the second stage (the superscript $u$ stands for \emph{unstructured}). This comparison highlights the value of the custom-designed kernel and quantifies the resulting performance gains.

For each of the four systems, we performed a Monte Carlo study with 2000 runs. In each run, the measurement noise $e$ in \eqref{eq:output-eq} varied, covering scenarios from very low to high noise. Accuracy of the estimated profile $\hat g$ relative to the true profile $g$ was quantified using the Fit measure:
\[
\text{Fit} = 100 \Bigg( 1 - \frac{\|g - \hat g\|}{\|g\|} \Bigg),
\]
where 100 indicates perfect reconstruction.

Table \ref{Table1} reports the average fits returned by Oracle, Trine, Trine$^u$ and MLH-GP. Detailed results for Toggle Switch and FitzHugh--Nagumo, the two most complex systems, are shown in Fig.~\ref{FigBoxplot}, which presents boxplots of the 2000 Fit values for each estimator. Trine clearly outperforms state-of-the-art approaches such as MLH-GP, demonstrating that modeling intrinsic noise with a smooth Gaussian prior is more effective than log-normal formulations, which are heavy-tailed, sensitive to measurement noise, and prone to overfitting. Additionally, the computational cost of Trine is much lower than that of MLH-GP, because the latter is an iterative method, whereas Trine provides solutions by solving only three subproblems. Even more importantly, Trine's performance is very close to that of the Oracle. As shown in Table~\ref{Table1}, the average Fit of the Oracle across the four systems was around $91\%$. In comparison, Trine achieved average Fits that were only a few percentage points below the Oracle, with at most a three percentage point difference across all systems. Comparison with Trine$^{\mathbf{u}}$ further emphasizes that achieving performance close to the Oracle critically depends on the structured kernel in Trine’s second stage. This kernel allows regularization and reliable estimation of intrinsic noise by capturing both the white-noise characteristics and the smooth state-dependent variance.

\begin{figure*}[h]
	\begin{center}
                \includegraphics[scale=0.7]{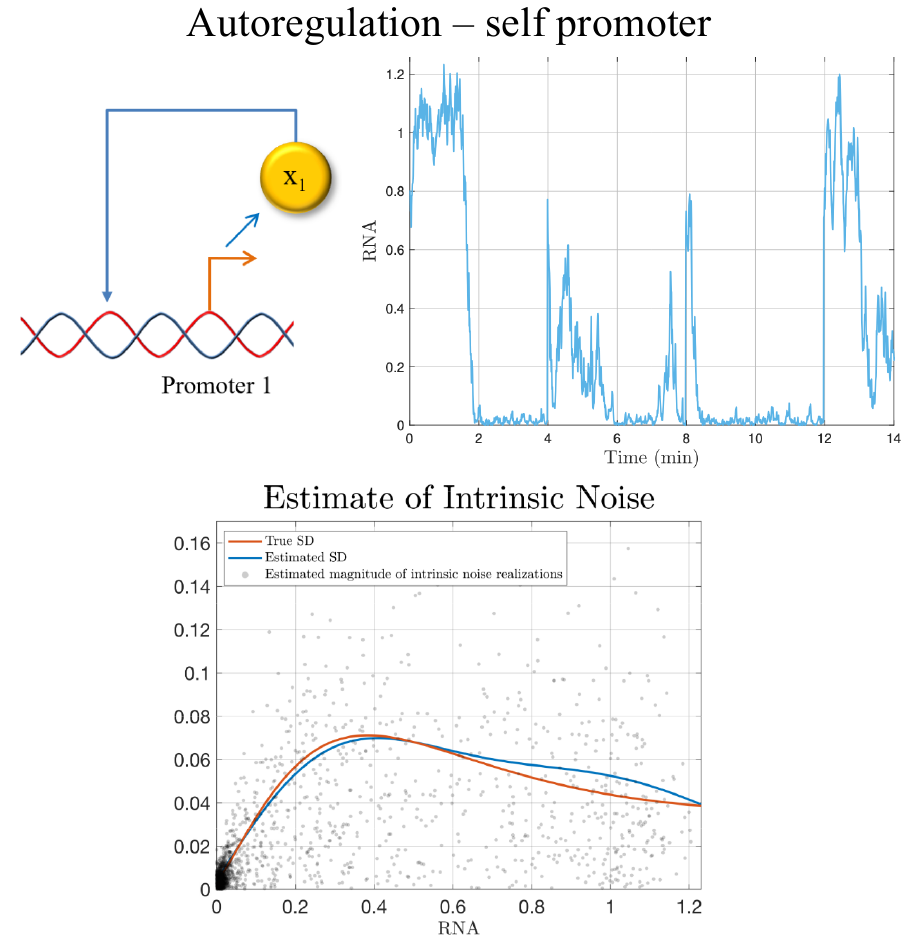}
		\caption{Results for Self-Promoter model. The top left panel shows the self-promoter gene regulatory network. The top right panel displays a representative simulated trajectory, whose dynamics exhibit pronounced burst-like stochastic behavior. Data are generated using the parameters $a_0 = 0.05$ (basal activity), $b = 10$ (feedback strength), $m_0 = 25$ (copy-number scale), $\kappa = 1$ (promoter-switching noise) \cite{Elston2001}.
        The bottom panel reports the true SD profile (red) and the Trine estimate (blue).}  \label{FigGene1D}
	\end{center}
\end{figure*}

\begin{figure*}[h]
	\begin{center}
        \includegraphics[scale=0.7]{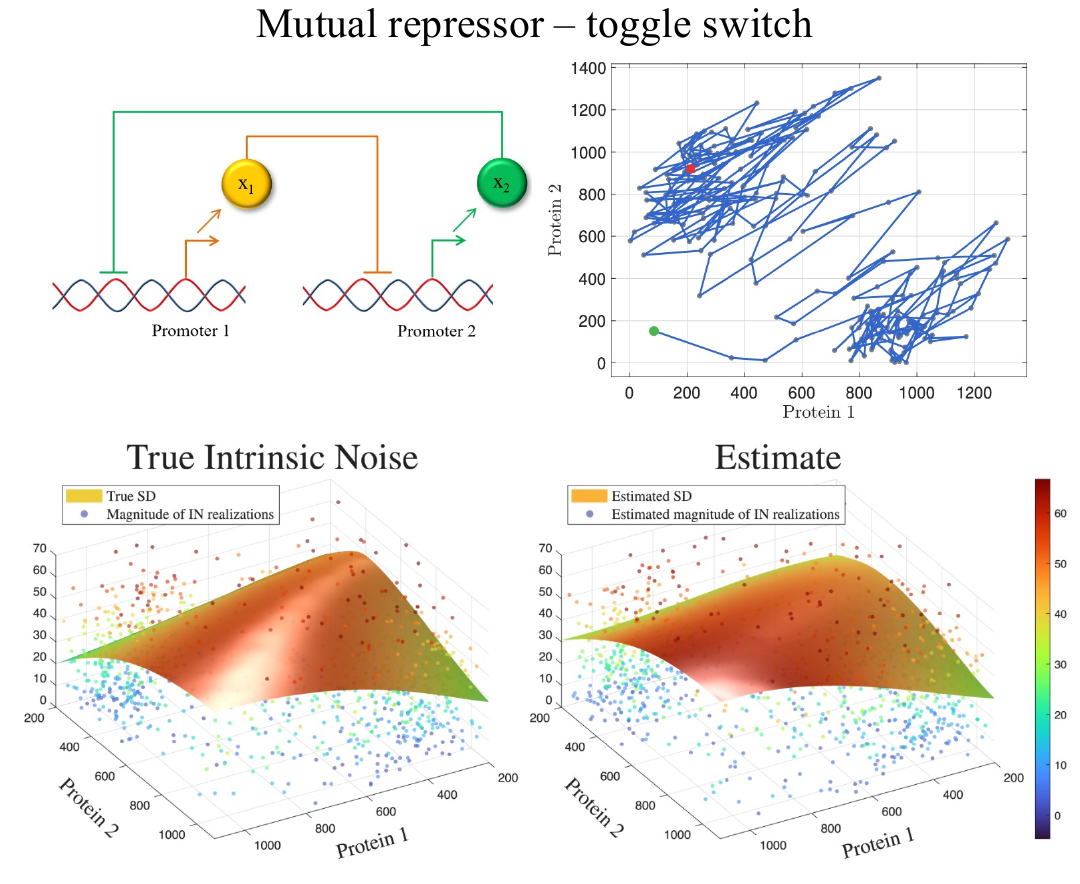}
	\end{center}
    \caption{Results for Toggle switch model. 
    The top left panel shows the gene regulatory network, while the top right panel shows a representative simulated trajectory of the two proteins in the 2D plane. Their stochastic dynamics show two basins of attraction, typical of bistable systems. The left-bottom panel shows the true intrinsic noise and state-dependent standard deviation while Trine estimates are in the right-bottom panel. Data are generated using the parameters $b = 0.28$, $m_0 = 1000$, $\kappa = 2.37$ (see Appendix for further details).}  \label{FigGene2D}
\end{figure*}

\begin{figure*}[h]
    \centering
    \makebox[\textwidth][c]{%
        \includegraphics[width=1.2\textwidth]{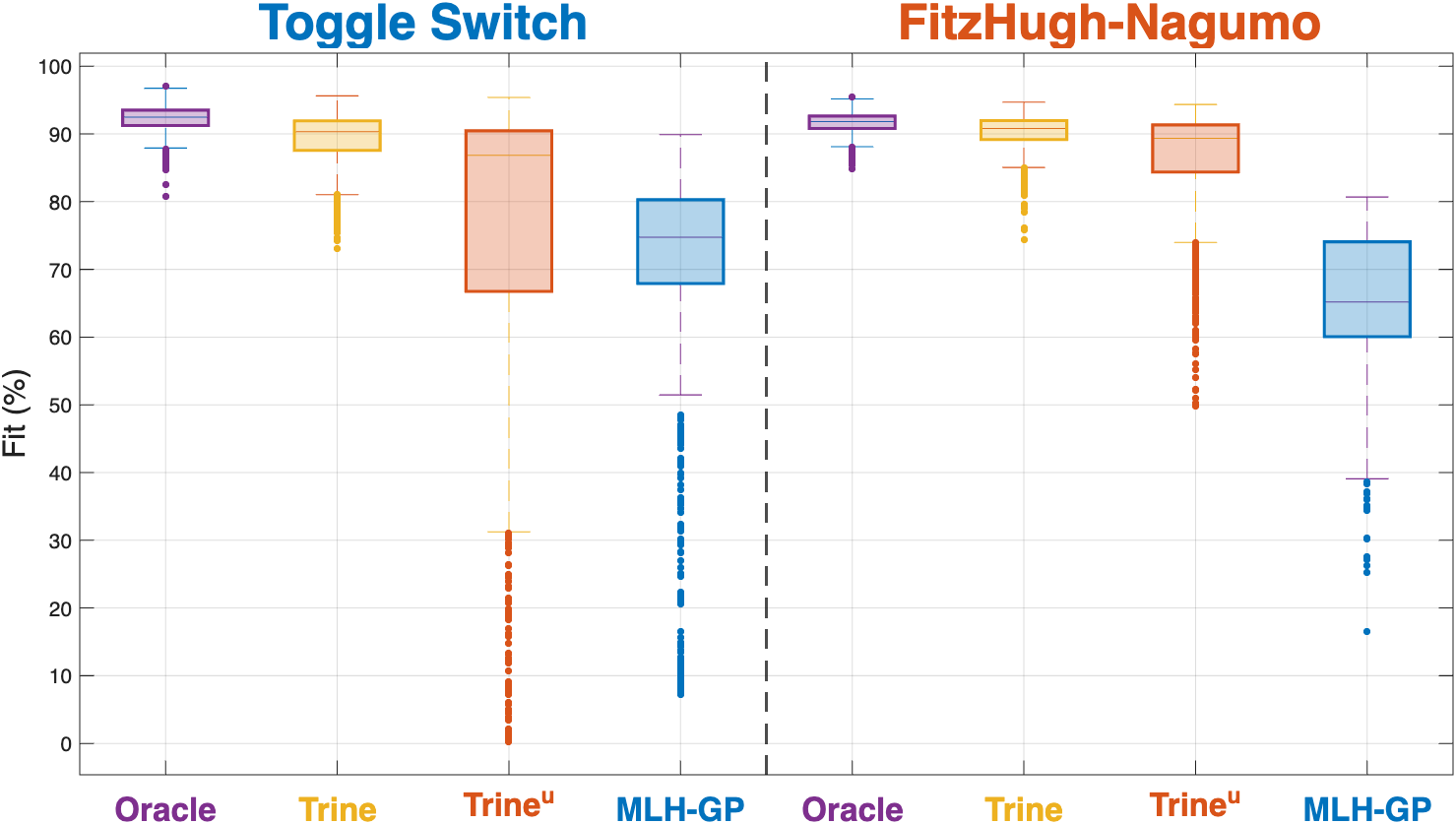}%
    }
		\caption{Boxplots of the Fit values (percentage accuracy) obtained from 2000 Monte Carlo runs for each estimator across the Toggle Switch (left block) and the FitzHugh--Nagumo system (right block). The experiment is designed such that, in each run, the observed noise level can vary. Specifically, the ratio between the norm of measurement and intrinsic noise may vary uniformly over [0,0.4]. The four estimators compared are Oracle, Trine, Trine$^{\mathbf{u}}$, and MLH-GP. The figure highlights that Trine performs very close to the Oracle, an ideal procedure which can directly measure intrinsic noise realizations, and substantially outperforms the classical MLH-GP approach \cite{Kersting2007}, which uses a log-normal prior to model the state-dependent variance. Comparison with Trine$^{\mathbf{u}}$, a variant in which the third stage of Trine relies directly on the intrinsic noise estimates from the first stage, emphasizes the importance of the structured kernel in regularizing and accurately estimating the intrinsic noise profile.}
        \label{FigBoxplot}
\end{figure*}

\begin{table}[h!]
\centering
\caption{Average Fit (\%) obtained by each estimator across the four examples}
\label{Table1}

\resizebox{\textwidth}{!}{
\begin{tabular}{lcccc}
\toprule
\textbf{Estimator} &
\textbf{Ricker} &
\textbf{Self promoter} &
\textbf{Toggle Switch} &
\textbf{FitzHugh–Nagumo} \\
\midrule
Oracle        & 93.6 & 88.1 & 92.2 & 91.6 \\
Trine         & 91.1 & 87.2 & 89.1 & 90.3 \\
Trine$^{u}$   & 82.2 & 78.1 & 72.6 & 79.3 \\
MLH-GP        & 78.7 & 61.8 & 67.1 & 61.4 \\
\bottomrule
\end{tabular}
}
\end{table}

\section*{Conclusions}

The Trine algorithm introduces a flexible yet rigorously designed framework for disentangling intrinsic noise in stochastic dynamical systems. By integrating nonparametric kernel methods within a modular three-phase design, Trine addresses the core challenge of estimating state-dependent noise profiles without restrictive parametric assumptions and remains robust even under limited or noisy data. Beyond accurate estimation, Trine provides a clear and interpretable view of the stochastic mechanisms shaping system dynamics, particularly in biological and ecological contexts, where noise is a meaningful functional component rather than a mere perturbation. This interpretability makes Trine not only a predictive tool but also a valuable instrument for discovery, revealing which molecular or dynamical variables modulate stochasticity and uncovering functional symmetries or nonlinear dependencies often obscured by experimental noise.

Estimating state-dependent variance is fundamental because it links statistical inference to biological meaning. Intrinsic noise is unavoidable in cellular systems. 
Quantifying how its variability depends on the system state provides a direct measure of robustness and sensitivity and allows both modelers and biologists to understand which regions of the state space are most influenced by stochastic fluctuations. Notably, recent control-theoretic advances in synthetic biology have shown that stochastic variability can itself be a design target, with feedback architectures capable of modulating or even reducing the stationary variance of biochemical networks (e.g., antithetic integral feedback \cite{Briat2018}). In this context, reconstructing the full state-dependent diffusion profile 
reveals natural targets for noise-aware control or intervention strategies.

To assess Trine’s contribution to the field, we conducted an extensive benchmarking analysis against other state-of-the-art approaches, in particular \cite{Kersting2007}, which serves as a strong baseline in this area. These tests demonstrated Trine’s superiority not only in reconstruction accuracy but also in robustness to measurement noise and uncertainty in system states.  
To further quantify this advancement, we introduced an ``oracle'' comparator---a theoretical benchmark representing the ideal upper bound of performance. In statistics, an oracle estimator has access to privileged information typically hidden in real experiments. Our oracle is envisioned as an idealized observer that can measure intrinsic noise realizations without error, obtaining information inaccessible in actual experiments. Remarkably, despite being entirely data-driven and lacking any privileged information, Trine achieves performance close to the oracle, narrowing the divide between theoretical stochastic modeling and experimental observability.  The structured kernel introduced in Trine’s second phase is essential for this performance, simultaneously enforcing smooth dependence of intrinsic noise variance on the state and modeling sharp discontinuities in its realizations.

Across benchmark systems, Trine consistently uncovered biologically meaningful stochastic structures. In the genetic toggle switch, it identified regions of the state space most sensitive to noise-driven transitions, illustrating how intrinsic fluctuations shape the landscape of cell-fate decisions. Similar insights were observed in the self-promoter, FitzHugh–Nagumo, and Ricker models, where Trine successfully recovered characteristic variance profiles despite strong nonlinearities and uneven sampling. Collectively, these four paradigmatic systems represent key classes of noisy biological dynamics---feedback regulation, bistability/excitability, oscillations, and stochastic population growth---highlighting both the biological significance and the broad applicability of Trine.

Looking ahead, several avenues exist for further development of the framework. One promising direction is integrating Trine with experimental design strategies to optimize data collection for noise estimation, enhancing both accuracy and efficiency. Additionally, coupling Trine with real-time inference or adaptive control methods could open new possibilities for monitoring and steering stochastic systems in both experimental and applied settings. Moreover, the principles underlying Trine extend far beyond biological and ecological modeling. Any domain where state-dependent noise is relevant—including neuroscience, financial modeling, and complex engineering systems—can benefit from this framework. 

\section*{Appendix} 

\subsection*{Trine algorithm}

To introduce Trine, it is first useful to establish some preliminary notation.
We define a delayed version of the measured states as
$$
z_k := y_{k+1}, \quad k = 1, \dots, N-1.
$$
This allows the dataset to be reformulated as a collection of $N-1$ training pairs $\{(y_k, z_k)\}_{k=1}^{N-1}$, where each pair corresponds to a fixed time shift. We assume that the time series $\{y_k\}$ is uniformly sampled, which ensures consistency in the temporal spacing
$\Delta_t$ across all training examples. This regularity is crucial for the algorithm to properly exploit the temporal structure of the data.
In continuous time, the following approximation, related to the Euler–Maruyama method \cite{kloeden1992numerical}, holds:
$$
x_{k+1} \approx x_k + \Delta_t f(x_k) + \sqrt{\Delta_t}\, g(x_k) w_k
$$
where $w_k$ are independent random variables with zero mean and unit variance.

It follows that, after training, Trine provides estimates for the deterministic part 
$F(x_k) := x_k + \Delta_t f(x_k)$ and the stochastic part $G(x_k) := \sqrt{\Delta_t}\, g(x_k)$. 
Once estimates of $F$ and $G$ are obtained, the corresponding $f$ and $g$ can be directly recovered.
The notation $F$ and $G$ is adopted in the pseudo-code reported below, 
but was not used in Fig. \ref{Fig1}, in order to avoid overloading the notation and keep the earlier presentation lighter.\\

It is also convenient to define the vector
$$
z = [z_1, \dots, z_{N-1}]^\top.
$$
In addition, given a kernel function $\mathcal{K}$, we define the associated \emph{kernel matrix} $\mathbb{K} \in \mathbb{R}^{(N-1) \times (N-1)}$ as
$$
\mathbb{K}_{ij} = \mathcal{K}(y_i, y_j).
$$
The kernel can also be enriched by including a so-called \emph{bias space} \cite{SpringerRegBook2022}, for example by adding constant or linear terms to capture trends in the data. These components are omitted here for notational simplicity, but their inclusion in the procedure is straightforward.\\
The pseudo-code of Trine is reported below. 
\bigskip
\bigskip

\noindent {\bf{Trine pseudocode}}
\medskip
\begin{algorithmic}[1]
\Statex \textbf{Input:} noisy states $y_k$ and covariance matrix $\Sigma_e$ of the output noise.
\Statex \textbf{Input hyperparameters used in the step:} 
\Statex \quad Step 1: $\lambda_f, \ell_f, \rho_n$
\Statex \quad Step 2: $\lambda_f, \ell_f,\lambda_w,\ell_w$
\Statex \quad Step 3: $\lambda_g, \ell_g, \rho_g$
\State \textbf{Step 1: Estimate intrinsic noise signs using a smooth kernel for the deterministic part $F$ and assuming intrinsic noise of constant variance}
\State Introduce the Gaussian kernel as
$$
\mathcal{K}_f(x,x') = \lambda_f \exp\left(-\frac{\|x - x'\|^2}{2 \ell_f}\right)
$$
\State Compute kernel matrix $\mathbb{K}_f$ using $\mathcal{K}_f$ evaluated on noisy states $y_k$
$$
\mathbb{K}_f \in \mathbb{R}^{(N-1) \times (N-1)}, \quad (\mathbb{K}_f)_{ij} = \mathcal{K}_f(y_i, y_j)
$$
\State Compute estimates of intrinsic noise signs:
$$
s = \mathrm{sign}\Big((\mathbb{K}_f + \Sigma_e + \rho_n I)^{-1} z\Big)
$$
\State Build the sign matrix:
$$
\mathbf{S} = \mathrm{diag}(s)
$$
\State \textbf{Step 2: Estimate the intrinsic noise realizations using sign-informed kernel}
\State Introduce the following Gaussian kernel matrix:
$$
\mathbf{G} \in \mathbb{R}^{(N-1) \times (N-1)}, \quad \mathbf{G}_{ij} = \lambda_w \exp\left(-\frac{\|y_i - y_j\|^2}{2 \ell_w}\right)
$$
\State Define the following matrix that will serve as a correction factor for $\mathbf{G}$:
$$
\mathbf{Q} =  \beta^2 \mathbf{1} \mathbf{1}^\top + \left(1 - \beta^2 \right) \mathbf{I}
$$
where $\mathbf{1} \in \mathbb{R}^{N-1}$ is the column vector of ones, $\mathbf{I}$ is the identity matrix
and $\beta=\sqrt{\frac{2}{\pi}}$ for the intrinsic Gaussian noise case.
\State Define the structured kernel matrix modeling intrinsic noise realizations as combination of $\mathbf{G},\mathbf{Q}$ and the sign matrix $\mathbf{S}$:
$$
\mathbb{K}_{gw}  = \mathbf{S} \cdot (\mathbf{G} \circ \mathbf{Q}) \cdot \mathbf{S}
$$
where $\circ$ denotes the Hadamard (element-wise) product.
\State Compute the estimates of the intrinsic noise realizations:
$$
\hat{n} = \mathbb{K}_{gw} \cdot (\mathbb{K}_f + \Sigma_e + \mathbb{K}_{gw})^{-1} \cdot z
$$
\State \textbf{Step 3: Estimate the intrinsic noise standard deviation}
\State Define the following Gaussian kernel matrix to model $G$:
$$
\mathbb{K}_g \in \mathbb{R}^{(N-1) \times (N-1)}, \quad (\mathbb{K}_g)_{ij} = \lambda_g \exp\left(-\frac{\|y_i - y_j\|^2}{2 \ell_g}\right)
$$
\State Define the weights
$$
\hat{c} = \left(\mathbb{K}_g + \rho_g I\right)^{-1} \cdot \frac{ |\hat{n}|}{\beta} 
$$
\State Compute the SD profile as:
$$
\widehat{G} = \mathbb{K}_g \hat{c} 
$$
\Statex \textbf{Output:} estimated intrinsic noise SD vector $\widehat{G}$ computed over the $y_k$ and weight vector $\hat{c}$ which, for any $x$, permits to calculate the SD estimate as follows
$$
\hat{G}(x) = \sum_{k=1}^{N-1}\hat{c}_k \lambda_g \exp\left(-\frac{\|x - y_k\|^2}{2 \ell_g}\right).
$$ 
\end{algorithmic}

\medskip

As explained in detail in Appendix, each stage of Trine estimates the unknown functions $f$, $g$, and $g \cdot w$ within the subspace spanned by the kernel sections $\mathcal{K}(y_i, \cdot)$, where $y_i$ are the observed (noisy) states. This construction, grounded in the representer theorem \cite{Wahba:90}, reflects the nonparametric nature of the method: the number of basis functions is not fixed in advance, but increases with the amount of data and can become infinite in the limit of dense sampling. The associated expansion coefficients are computed by solving regularized quadratic optimization problems, where the regularization is implicitly defined by the choice of kernel \cite{Scholkopf01b}.

The initial part of the pseudo-code specifies the hyperparameters adopted at each stage of the algorithm. These parameters are essential for regularization, as they govern the trade-off between fidelity to the empirical data and adherence to the structural priors encoded in the kernel. A distinctive feature of the Trine methodology is its systematic formulation of 
linear estimation subproblems. This formulation permits the closed-form evaluation of model selection criteria such as the Prediction Error Sum of Squares (PRESS), the Generalized Cross-Validation (GCV) score~\cite{WahbaG:79} or empirical Bayes methods \cite{SpringerRegBook2022}, thereby enabling computationally efficient and theoretically principled hyperparameter optimization. Specifically, in the present work, hyperparameters are consistently estimated via maximization of the marginal likelihood---also referred to as Bayesian evidence---a widely adopted approach in probabilistic modeling. This criterion quantifies the explanatory adequacy of the model with respect to the observed data while inherently incorporating the Occam’s razor principle~\cite{MacKayNC92}. 

The parameter $\beta$ is used in specific steps of the algorithm to define certain matrices and vectors. The value 
 $\sqrt{2/\pi}$ is calibrated for the case of Gaussian intrinsic noise. As detailed in Appendix, the constant can be modified if the noise is known to follow different statistical distributions, such as Laplacian or Student's $t$-distributions.

The final stage of the algorithm produces an unconstrained estimate of the diffusion component $g$. In our experiments, we did not impose any additional constraints. However, since the optimization problem in Step 3, which estimates the intrinsic noise variance, is quadratic, the framework can easily be extended to include structural constraints on $g$, such as nonnegativity, concavity, or monotonicity. These properties can be enforced by applying inequality constraints to discrete-time derivatives of any order, resulting in convex optimization problems that standard numerical solvers can efficiently handle. 

The next section provides a detailed step-by-step derivation of the algorithm, along with further theoretical insights into its performance and practical aspects.

\subsection*{Mathematical Derivation of the Trine Algorithm}

For completeness, the pseudocode for Trine is reproduced below. It matches the version provided in Materials and Methods.

\medskip

\noindent {\bf{\large Trine algorithm}}
\begin{algorithmic}[1]
\Statex \textbf{Input:} noisy states $y_k$ and covariance matrix $\Sigma_e$ of the output noise
\Statex \textbf{Input hyperparameters used in the step:} 
\Statex \quad Step 1: $\lambda_f, \ell_f, \rho_n$
\Statex \quad Step 2: $\lambda_f, \ell_f,\lambda_w,\ell_w$
\Statex \quad Step 3: $\lambda_g, \ell_g, \rho_g$
\State \textbf{Step 1: Estimate intrinsic noise signs using a smooth kernel for the deterministic part $F$ and assuming intrinsic noise of constant variance}
\State Introduce the Gaussian kernel as
$$
\mathcal{K}_f(x,x') = \lambda_f \exp\left(-\frac{\|x - x'\|^2}{2 \ell_f}\right)
$$
\State Compute kernel matrix $\mathbb{K}_f$ using $\mathcal{K}_f$ evaluated on noisy states $y_k$
$$
\mathbb{K}_f \in \mathbb{R}^{(N-1) \times (N-1)}, \quad (\mathbb{K}_f)_{ij} = \mathcal{K}_f(y_i, y_j)
$$
\State Compute estimates of intrinsic noise signs:
$$
s = \mathrm{sign}\Big((\mathbb{K}_f + \Sigma_e + \rho_n I)^{-1} z\Big)
$$
\State Build the sign matrix:
$$
\mathbf{S} = \mathrm{diag}(s)
$$
\State \textbf{Step 2: Estimate the intrinsic noise realizations using sign-informed kernel}
\State Introduce the following Gaussian kernel matrix:
$$
\mathbf{G} \in \mathbb{R}^{(N-1) \times (N-1)}, \quad \mathbf{G}_{ij} = \lambda_w \exp\left(-\frac{\|y_i - y_j\|^2}{2 \ell_w}\right)
$$
\State Define the following matrix that will serve as a correction factor for $\mathbf{G}$:
$$
\mathbf{Q} =  \beta^2 \mathbf{1} \mathbf{1}^\top + \left(1 - \beta^2 \right) \mathbf{I}
$$
where $\mathbf{1} \in \mathbb{R}^{N-1}$ is the column vector of ones, $\beta=\mathbb{E}[|w_k|]$ and $\mathbf{I}$
is the identity matrix.
\State Define the structured kernel matrix modeling intrinsic noise realizations as combination of $\mathbf{G},\mathbf{Q}$ and the sign matrix $\mathbf{S}$:
$$
\mathbb{K}_{gw}  = \mathbf{S} \cdot (\mathbf{G} \circ \mathbf{Q}) \cdot \mathbf{S}
$$
where $\circ$ denotes the Hadamard (element-wise) product.
\State Compute the estimates of the intrinsic noise realizations:
$$
\hat{n} = \mathbb{K}_{gw} \cdot (\mathbb{K}_f + \Sigma_e + \mathbb{K}_{gw})^{-1} \cdot z
$$
\State \textbf{Step 3: Estimate the intrinsic noise standard deviation}
\State Define the following Gaussian kernel matrix to model $G$:
$$
\mathbb{K}_g \in \mathbb{R}^{(N-1) \times (N-1)}, \quad (\mathbb{K}_g)_{ij} = \lambda_g \exp\left(-\frac{\|y_i - y_j\|^2}{2 \ell_g}\right)
$$
\State Define the weights
$$
\hat{c} = \left(\mathbb{K}_g + \rho_g I\right)^{-1} \cdot \frac{ |\hat{n}|}{\beta} 
$$
\State Compute the SD profile as:
$$
\widehat{G} = \mathbb{K}_g \hat{c} 
$$
\Statex \textbf{Output:} estimated intrinsic noise SD vector $\widehat{G}$ computed over the $y_k$ and weight vector $\hat{c}$ which, for any $x$, permits to calculate
$$
\hat{G}(x) = \sum_{k=1}^{N-1}\hat{c}_k \lambda_g \exp\left(-\frac{\|x - y_k\|^2}{2 \ell_g}\right).
$$ 
\end{algorithmic}

\medskip

The following sections are devoted to deriving the three stages of Trine through regularization theory in Reproducing Kernel Hilbert Spaces (RKHS), alongside its Bayesian interpretation.

\subsubsection*{Stage 1: Sign Estimation of the Intrinsic Noise}

We begin by reformulating the problem in a supervised learning framework. The available input–output training data consist of delayed state pairs:
$$
z_k := y_{k+1}, \quad \text{with training pairs } (y_k, z_k), \quad k = 1, \dots, N-1.
$$
This formulation allows us to learn a map from the observed inputs $y_k$ (which approximate the true states $x_k$) to the corresponding outputs $z_k$.

In particular, we consider the continuous-time system dynamics and approximate them in discrete time over the measurement sampling interval $\Delta_t$:  
$$
x_{k+1} \approx x_k + \Delta_t f(x_k) + \sqrt{\Delta_t}\, g(x_k) w_k,
$$
where $\Delta_t$ denotes the sampling period at which the noisy states $z_k$ are observed.  
We then define the functions
\begin{equation*}
F(x) := x + \Delta_t f(x), \qquad 
G(x) := \sqrt{\Delta_t}\, g(x),
\end{equation*}
so that the model can be written as
\begin{equation}\label{MeasMod}
z_k \approx F(x_k) + G(x_k) w_k + e_k.
\end{equation}

To estimate the signs of the intrinsic noise components $w_k$, which enter additively in the dynamics, we introduce two kernel models for $F$ and $G$, and perform regularized regression in Reproducing Kernel Hilbert Spaces (RKHSs). We assume only that $F$ is smooth, and associate to it a Gaussian kernel:
$$
\mathcal{K}_f(x,x') = \lambda_f \exp\left(-\frac{\|x - x'\|^2}{2 \ell_f^2}\right).
$$
Given that we only observe noisy approximations of the true states, the entries $x_k$ of $F,G$ in \eqref{MeasMod} are replaced
with $y_k$. Using regularization in RKHS, inspired by the representer theorem~\cite{Wahba:90} we assume that $F$ lies in the subspace of the RKHS $\mathcal{H}$ spanned by the kernel sections $\mathcal{K}(y_i, \cdot)$.
Specifically, we write
$$
F(\cdot) = \sum_{i=1}^{N-1} \alpha_i \mathcal{K}_f(y_i, \cdot),
$$
which implies that the squared RKHS norm of $F$, which serves as a regularization term, is given by
$$
\| F \|^2_{\mathcal{H}} = \alpha^\top \mathbb{K}_f \alpha,
$$
where $\mathbb{K}_f$ is the kernel matrix defined as
$$
(\mathbb{K}_f)_{ij} = \mathcal{K}_f(y_i, y_j), \quad \mathbb{K}_f \in \mathbb{R}^{(N-1) \times (N-1)},
$$
e.g., see \cite{Scholkopf01b}.
As for the stochastic component, at this stage of the algorithm we lack sufficient information to define a structured kernel. Since our goal here is only to estimate the intrinsic noise signs, the Bayesian interpretation of regularization in RKHSs---where kernels represent the covariance functions of Gaussian processes---suggests using a diagonal kernel with constant variance $\rho_n$.
This choice simply reflects the white noise nature of the intrinsic noise; it will be the task of the subsequent stages to refine $\rho_n$ by deriving a functional relationship between the variance and the state $x$.\\
By the representer theorem, the estimate of the intrinsic noise within the RKHS can also be written as a linear combination of kernel functions centered at the data points. 
In this specific case, since the kernel is diagonal with constant value $\rho_n$, the estimate reduces simply to the product between the scalar $\rho_n$ and the unknown coefficient vector $c \in \mathbb{R}^{N-1}$. 
Hence, using \eqref{MeasMod}, we can express the measurement model as
\begin{equation}
z = \mathbb{K}_f \alpha + \rho_n c + E,
\end{equation}
where the $(N-1)$-dimensional random vector $E$ collects the noise terms $e_k$ and has covariance $\Sigma_e$. 
The vectors $\alpha$ and $c$ are unknown coefficients, whose estimation enables the recovery of the function $F$ and the intrinsic noise, respectively. 
Their regularized estimates are obtained by solving the optimization problem
$$
(\hat{\alpha}, \hat{c}) = \arg\min_{\alpha,c} \; (z - \mathbb{K}_f \alpha - \rho_n c)^\top \Sigma_e^{-1} (z - \mathbb{K}_f \alpha - \rho_n c) + \alpha^\top \mathbb{K}_f \alpha + \rho_n c^\top c.
$$
One has
$$
\hat{\alpha} = \hat{c} = (\mathbb{K}_f + \Sigma_e + \rho_n I)^{-1} z,
$$
which correspond to the weights of a regularization network \cite{Scholkopf01b}.
In fact, the optimality conditions are
$$
\mathbb{K}_f \Sigma_e^{-1} (z - \mathbb{K}_f \hat{\alpha} - \rho_n \hat{c}) = \mathbb{K}_f \hat{\alpha},
$$
$$
\rho_n \Sigma_e^{-1} (z - \mathbb{K}_f \hat{\alpha} - \rho_n \hat{c}) = \rho_n \hat{c}.
$$
Substituting $\hat{\alpha} = \hat{c}$, both reduce to
$$
(\mathbb{K}_f + \rho_n I) \Sigma_e^{-1} (z - (\mathbb{K}_f + \rho_n I) \hat{c}) = (\mathbb{K}_f + \rho_n I) \hat{c}.
$$
and multiplying both sides by $(\mathbb{K}_f + \rho_n I)^{-1}$ and then by $\Sigma_e$, we obtain
$$
(\mathbb{K}_f + \Sigma_e + \rho_n I) \hat{c} = z,
$$
which is satisfied by the definition of $\hat{c}$.\\
The estimates of the intrinsic noise realizations are thus given by
\begin{equation}\label{EstINFirstStage}
\rho_n (\mathbb{K}_f + \Sigma_e + \rho_n I)^{-1} z,
\end{equation}
which explains the structure of the sign estimator appearing on line 4 of the Trine pseudocode.

\subsubsection*{Stage 2: Intrinsic Noise Estimation using a structured kernel}

The sign estimates obtained in Stage~1 are now employed to construct a structured kernel that models the realizations of the intrinsic noise. 
The incorporation of sign information is crucial in order to obtain a non-diagonal kernel to describe the intrinsic noise standard deviation, allowing us to capture its smooth dependence on the state value.\\
Recall that $G(x) = \sqrt{\Delta_t} \, g(x)$, so that the transition noise between time steps $k$ and $k+1$ can be expressed as:
$$
\mathrm{sign}(w_k) \, G(x_k) \, |w_k|,
$$
where $\mathrm{sign}(w_k)$ is set to the estimate obtained in the previous stage.

To capture the smooth variability of the function $G$, we assign it a Gaussian kernel:
$$
\mathcal{K}_g(x_a, x_b) = \lambda_w \exp\left(-\frac{\|x_a - x_b\|^2}{2 \ell_w}\right).
$$

To obtain the overall kernel
for the intrinsic noise, we exploit the Bayesian interpretation of regularization in RKHS. The function $G$ is seen as a zero-mean Gaussian random field of covariance $\mathcal{K}_g$, independent of $w_k$. Hence, the covariance of the random process 
$\mathrm{sign}(w) \, G(x) \, |w|$, under the assumption of known signs, is
$$
\mathrm{Cov}\left( \mathrm{sign}(w_a) G(x_a) |w_a|, \mathrm{sign}(w_b) G(x_b) |w_b| \right)
= 
\begin{cases}
\mathcal{K}_g(x_a,x_a) & \text{if } a = b \\
\mathrm{sign}(w_a)\mathrm{sign}(w_b) \, \mathcal{K}_g(x_a,x_b) \, \beta^2 & \text{if } a \neq b
\end{cases}
$$

Here, $\beta = \mathbb{E}[|w_k|]$ captures the expected magnitude of the intrinsic noise. Note that for standard normal noise, $\beta = \sqrt{2/\pi}$.

Following the same principle as in Stage 1, we assume that the intrinsic noise realizations lie in the subspace spanned by the kernel sections centered at the observed noisy states $y_k$. In practice, this means that the input locations used for learning are the noisy estimates $y_k$, rather than the true but unobservable states $x_k$. From the covariance expression above, we recover the structured kernel matrix $\mathbb{K}_{gw}$ used in the pseudocode (line 9), which incorporates both the Gaussian kernel and the sign information. The deterministic part $F$ is still assigned the Gaussian kernel $\mathcal{K}_f$. 
This completes our kernel model and, recalling \eqref{MeasMod}, it follows that:
$$
z = \mathbb{K}_f \alpha + \mathbb{K}_{gw} c + E,
$$
where $E$ is the vector of output noise.

We estimate $\alpha$ and $c$ by solving the regularized least squares problem:
$$
(\hat{\alpha}, \hat{c}) = \arg\min_{\alpha,c} \; (z - \mathbb{K}_f \alpha - \mathbb{K}_{gw} c)^\top \Sigma_e^{-1} (z - \mathbb{K}_f \alpha - \mathbb{K}_{gw} c) + \alpha^\top \mathbb{K}_f \alpha + c^\top \mathbb{K}_{gw} c,
$$
which has the closed-form solution:
$$
\hat{\alpha} = \hat{c} = (\mathbb{K}_f + \Sigma_e + \mathbb{K}_{gw})^{-1} z.
$$

This leads to the estimate of the intrinsic noise realizations:
\begin{equation}\label{EstINSecStage}
\hat{n} = \mathbb{K}_{gw} \cdot (\mathbb{K}_f + \Sigma_e + \mathbb{K}_{gw})^{-1} \cdot z,
\end{equation}
as given in line 10 of the Trine pseudocode.

Finally, one may optionally include a constant mean term $\mu$ in the model of $G(x)$. Since $\mathrm{sign}(w_k)$ is assumed  known, the process $\mathrm{sign}(w_k) G(x_k) |w_k|$ has a mean vector given by $s \circ (\mu \beta)$, where $s = [\mathrm{sign}(w_1), \ldots, \mathrm{sign}(w_{N-1})]^\top$,  the symbol $\circ$ denotes the Hadamard (element-wise) product and $\beta = \mathbb{E}[|w_k|]$. The kernel matrix $\mathbb{K}_{gw}$ is corrected by an additional diagonal term $\mu^2 (1 - \beta^2) \mathbf{I}$ to account for the increased variance. 
The estimate then becomes:
$$
\hat{n} = s \circ (\mu \beta) + \left(\mathbb{K}_{gw} + \mu^2 (1 - \beta^2) \mathbf{I}\right) \left(\mathbb{K}_f + \Sigma_e + \mathbb{K}_{gw} + \mu^2 (1 - \beta^2) \mathbf{I}\right)^{-1} \left(z - s \circ (\mu \beta)\right).
$$

\subsubsection*{Stage 3: Intrinsic Noise Variance Estimation}

At this stage, we aim to estimate the profile of the intrinsic noise standard deviation function, denoted $G(x)$. To motivate the estimation strategy, consider that for a fixed value of $x(t)$, the random variable $G(x)|w_t|$ represents the absolute value of the intrinsic noise component at that point. Since the noise variable $w_t$ has zero mean and unit variance, its absolute value has expected value $\beta = \mathbb{E}[|w_t|]$. Therefore, the expectation of $G(x)|w_t|$ is simply $G(x)\beta$.

From Stage 2 of the algorithm, we obtain the vector $\hat{n}$ reported in \eqref{EstINSecStage}, which estimates the realizations of the intrinsic noise. By taking its absolute value and dividing by $\beta$, we construct the vector $|\hat{n}|/\beta$, which can be interpreted as a pointwise estimate of the standard deviation profile $G$ at the noisy states $y_k$.

To recover a smooth function $G(x)$, we perform a regularized kernel regression. We associate to $G$ the Gaussian kernel $\mathcal{K}_g$ and use the corresponding kernel matrix $\mathbb{K}_g \in \mathbb{R}^{(N-1) \times (N-1)}$, as defined in line 12 of the Trine pseudocode. Given the regularization parameter $\rho_g$, we solve the following regularized least squares problem:
$$
\hat{c} = \arg\min_{c} \left\| \frac{|\hat{n}|}{\beta} - \mathbb{K}_g c \right\|^2 + \rho_g c^\top \mathbb{K}_g c
$$
This yields the closed-form solution:
$$
\hat{c} = \left(\mathbb{K}_g + \rho_g I\right)^{-1} \cdot \frac{|\hat{n}|}{\beta}
$$
and the estimates of the standard deviations at the $y_k$ are contained in the vector
\begin{equation}\label{widehatG}
    \widehat{G} = \mathbb{K}_g \hat{c}. 
\end{equation}
The estimate at a generic $x$ is then given by the sum of kernel sections centred at the $y_k$:
$$
\hat{G}(x) = \sum_{k=1}^{N-1} \hat{c}_k \lambda_g \exp\left(-\frac{\|x - y_k\|^2}{2 \ell_g}\right)
$$
The three expressions above correspond to those reported in lines 13 and 14 of the Trine pseudocode.

Since the entries of $|\hat{n}|/\beta$ are nonnegative by construction, the components of $\widehat{G}$ are typically nonnegative. Indeed, in all simulations presented in the main text, we observed that such estimated profile did not require explicit constraints.
Nevertheless, since the regression problem is a convex quadratic program, it is straightforward to incorporate additional linear constraints if desired. In particular, any linear constraint on the vector $q = \mathcal{K}_g \hat{c}$—which corresponds to the estimated values of $G$ at the training points $y_k$—can be added while preserving convexity. For example, enforcing the positivity of first-order discrete differences of $q$ promotes monotonicity, while requiring the negativity of second-order differences enforces concavity. These constraints can encode soft, high-level prior knowledge provided by the user, and can be seamlessly incorporated into the estimation process using standard convex optimization techniques. 
Constraints can also be imposed on locations different from the training points $y_k$ by extending the definition of the estimated profile $q$ to an arbitrary evaluation grid. In this setting, $q$ is expressed as $q = \mathcal{K}_{\text{grid}} c$,
where $\mathcal{K}_{\text{grid}}$ is the kernel matrix evaluating the Gaussian kernel between the evaluation grid points and the original data points. To ensure consistency with the observed data, a selection matrix $M$ is introduced, which extracts the components of $q$ corresponding to the training points $y_k$. Linear constraints can then be applied directly to the extended vector $q$, while the resulting optimization problem remains a convex quadratic program that can be efficiently solved using standard methods.\\
\medskip

Finally, recall that in real applications the hyperparameters 
$\lambda_f, \ell_f, \rho_n$ in the first stage, 
$\lambda_f, \ell_f, \lambda_w, \ell_w$ in the second stage, 
and $\lambda_g, \ell_g, \rho_g$ in the third stage 
are unknown. Since at each step of Trine the problem reduces to regularized least squares in RKHS, 
standard criteria such as GCV, PRESS, or marginal likelihood optimization can be employed for their estimation~\cite{Tibshirani2001,SpringerRegBook2022}. 
In all the experiments, the latter approach has been adopted, as implemented in the 
\texttt{Statistics and Machine Learning Toolbox} for Matlab. 


\subsection*{Intrinsic noise estimation: role of sign knowledge}

In this section, we investigate how intrinsic noise estimation may benefit from knowing the signs of the intrinsic noise, as estimated in the first stage of Trine. Theoretical findings will also help interpret more effectively the results from a Monte Carlo study of the toggle switch example, presented in the next section.\\

Consider a simplified setting where the drift $f$ is either absent or assumed to be known.
The intrinsic noise standard deviation is modeled as the discrete-time stochastic process 
$g_k + \mu$, indexed by $k = 1, 2, \ldots$. Here, $\mu$ is a positive scalar, and first- and 
second-order moments of $g_k$ follows those of the first-order autoregressive process (AR(1)):
$$
g_{k+1} = a g_k + v_k, \quad k = 1, 2, \ldots
$$
where:
\begin{itemize}
\item $v_k$ is white noise (all the random variables are mutually independent) with zero mean and variance $\ell^2$;
\item $0 < a < 1$, so that for large $k$, the process $g_k$ reaches stationarity with zero mean and variance
$$
\mathrm{Var}(g_k) = \gamma^2 := \frac{\ell^2}{1 - a^2}, \quad \mathrm{Cov}(g_k, g_j) = \gamma^2 a^{|k-j|}.
$$
\end{itemize}
The AR(1) process is simple yet significant. Its parameter $a$ plays a role similar to the kernel width in a Gaussian kernel, as it determines the correlation between samples, thus controlling the smoothness of the intrinsic noise SD profile. As shown below, this structure allows for the derivation of closed-form expressions for the estimation performance.\\

The intrinsic noise is then given by $(g_k + \mu)w_k$, where $w_k$ is a white noise process, independent of $g_k$, with zero mean and unit variance:
$$
\mathbb{E}[w_k] = 0, \quad \mathbb{E}[w_k^2] = 1, \quad \mathbb{E}[|w_k|] := \beta \ge 0.
$$

We consider two possible scenarios for the estimation task. In the first scenario, the signs of the intrinsic noise are unknown, and we define:
$$
z^u_k = (g_k + \mu)w_k, \quad k = 1, 2, \ldots.
$$
The measured output process $y^u_k$ is:
$$
y^u_k = z^u_k + e_k, \quad k = 1, 2, \ldots
$$
where $e_k$ is white noise with variance $\sigma^2$.\\

In the second scenario, the signs are assumed to be known:
$$
z_k = (g_k + \mu)|w_k|, \quad k = 1, 2, \ldots
$$
and
$$
y_k = z_k + e_k, \quad k = 1, 2, \ldots
$$
The processes $g_k$, $w_k$, and $e_k$ are all mutually independent.

Our goal is to compute and compare the mean square error (MSE) of the optimal linear estimators of $z^u_k$ and $z_k$, based on all output measurements available up to time $k$, in the steady-state regime (i.e., for large $k$). Note that the case with known signs corresponds to 
the second stage of Trine, and that accurate estimates of $z^u$ or $z$ (here representing the intrinsic noise) are crucial for the successful implementation of Trine third stage (which estimates the variance profile exploiting the intrinsic noise estimates).

\subsubsection*{Estimation of $z^u_k$}

For the stochastic processes
$$
z^u_k = (g_k + \mu)w_k, \quad y^u_k = z^u_k + e_k,
$$
for large $k$, simple considerations exploiting the independence among $g$, $w$, and $e$, as well as the whiteness of $w$ and $e$, lead to:
$$
\mathbb{E}[z^u_k] = 0,
$$
$$
\mathrm{Cov}(z^u_k, y^u_k) = \mathrm{Var}(z^u_k) = \gamma^2 + \mu^2,
$$
$$
\mathrm{Var}(y^u_k) = \gamma^2 + \mu^2 + \sigma^2,
$$
and
$$
\mathrm{Cov}(z^u_k, y^u_j) = \mathrm{Cov}(z^u_k, z^u_j) = 0, \quad \text{for } k \neq j.
$$

It follows that the optimal linear estimator of $z^u_k$ depends solely on $y^u_k$, since all previous measurements $\{y^u_i\}_{i=1}^{k-1}$ are uncorrelated with $z^u_k$  \cite{Anderson:1979}. This implies that the estimator cannot exploit any correlation or smoothness structure in the standard deviation profile.\\

The optimal estimator is therefore:
$$
\hat{z}^u_k =
\frac{\mathrm{Cov}(z^u_k, y^u_k)}
{\mathrm{Var}(y^u_k)}
y^u_k = \frac{\gamma^2 + \mu^2}{ \gamma^2 + \mu^2 + \sigma^2} y^u_k.
$$

The MSE of this estimator is
$$
\mathrm{MSE}_1 = \mathrm{Var}(z^u_k - \hat{z}^u_k)
= \mathrm{Var}(z^u_k)-
\frac{\mathrm{Cov}(z^u_k, y^u_k)^2}
{\mathrm{Var}(y^u_k)}
= (\gamma^2 + \mu^2)- \frac{(\gamma^2 + \mu^2)^2}
{\gamma^2 + \mu^2 + \sigma^2}.
$$

This expression can be equivalently rewritten as:
$$
\mathrm{MSE}_1 =
\frac{(\gamma^2 + \mu^2)\sigma^2}{\gamma^2 + \mu^2 + \sigma^2}.
$$

\subsubsection*{Estimation of $z_k$}

The derivation of the MSE for the linear estimator of $z_k$, based on the set $\{y_i\}_{i=1}^k$, is significantly more involved. Some preliminary steps are necessary. The key idea is to construct a state-space model that shares the same first- and second-order moments as $z_k$ and $y_k$. We then apply Kalman filtering techniques to derive a closed-form expression for the asymptotic filtered covariance, which corresponds to the desired MSE \cite{Anderson:1979}.\\

\subsubsection*{First- and second-order moments of $z_k$}

Simple calculations yield the following expressions for the mean and variance of $z_k$:
\begin{eqnarray*}
\mathbb{E}[z_k] &=& \mathbb{E}[(g_k + \mu)|w_k|] \\
&=& \mathbb{E}[g_k] \cdot \mathbb{E}[|w_k|] + \mu \cdot \mathbb{E}[|w_k|] \quad (\text{independence})\\
&=& \mu \beta,
\end{eqnarray*}

\begin{eqnarray*}
\mathrm{Var}(z_k) &=& \mathbb{E}[z_k^2] - (\mathbb{E}[z_k])^2 \\
&=& \mathbb{E}[(g_k + \mu)^2] \cdot \mathbb{E}[w_k^2] - \mu^2 \beta^2 \\
&=& (\gamma^2 + \mu^2) - \mu^2 \beta^2 \\
&=& \gamma^2 + \mu^2(1 - \beta^2).
\end{eqnarray*}

The covariance between $z_k$ and $z_j$ for $k \neq j$ is given by:
\begin{eqnarray*}
\mathrm{Cov}(z_k, z_j) &=& \mathbb{E}[z_k z_j] - \mathbb{E}[z_k]\mathbb{E}[z_j] \\
&=& (\mathbb{E}[g_k g_j] + \mu^2) \cdot \beta^2 - \mu^2 \beta^2 \\
&=& (\mathrm{Cov}(g_k, g_j) + \mu^2)\beta^2 - \mu^2 \beta^2 \\
&=& \beta^2 \gamma^2 a^{|k - j|}.
\end{eqnarray*}

\subsubsection*{Two stochastic processes with the same moments as $z_k$ and $y_k$}

Consider the process
\begin{equation}\label{tildez}
\tilde{z}_k = m + q_k + \varepsilon_k,
\end{equation}
where:

\begin{itemize}
\item $m = \mu \beta$ is a constant mean term;
\item $q_k$ is an AR(1) process:
$$
q_{k+1} = a q_k + \eta_k,
$$
with innovation variance
$$
\mathrm{Var}(\eta_k) = \sigma_\eta^2 = \beta^2 \ell^2,
$$
so that the stationary variance of $q_k$ is
$$
\mathrm{Var}(q_k) = \beta^2 \gamma^2;
$$
\item $\varepsilon_k$ is white noise, independent of $q_k$, with variance
$$
\sigma_\varepsilon^2 = (\gamma^2 + \mu^2)(1 - \beta^2).
$$
\end{itemize}

We now show that $\tilde{z}_k$ matches the first- and second-order moments of $z_k$.

Since both $q_k$ and $\varepsilon_k$ are zero-mean, we have:
$$
\mathbb{E}[\tilde{z}_k] = m = \mu \beta = \mathbb{E}[z_k].
$$

For the covariance between distinct time indices $k \ne j$:
$$
\mathrm{Cov}(\tilde{z}_k, \tilde{z}_j) = \mathrm{Cov}(q_k, q_j) = \beta^2 \gamma^2 a^{|k - j|} = \mathrm{Cov}(z_k, z_j).
$$

Finally, for the variance:
$$
\begin{aligned}
\mathrm{Var}(\tilde{z}_k) &= \mathrm{Var}(q_k) + \mathrm{Var}(\varepsilon_k) \\
&= \beta^2 \gamma^2 + (\gamma^2 + \mu^2)(1 - \beta^2) \\
&= \gamma^2 + \mu^2(1 - \beta^2) \\
&= \mathrm{Var}(z_k).
\end{aligned}
$$

We also define the corresponding output process as
$$
\tilde{y}_k = m + q_k + n_k,
$$
where
$$
m = \mu \beta, \quad n_k = \varepsilon_k + e_k.
$$
The process $n_k$ is white and independent of $q_k$, with variance
$$
\begin{aligned}
\mathrm{Var}(n_k) &= \sigma_n^2 \\
&= \mathrm{Var}(\varepsilon_k) + \mathrm{Var}(e_k) \\
&= (\gamma^2 + \mu^2)(1 - \beta^2) + \sigma^2.
\end{aligned}
$$

It follows that the process $\tilde{y}_k$ has the same first- and second-order moments as the original process $y_k$ defined from $z_k$. Moreover, the cross-covariances between the stochastic processes $\tilde{z}_k$ and $\tilde{y}_k$ exactly match those between $z_k$ and $y_k$. Therefore, the processes $\tilde{z}_k$ and $\tilde{y}_k$ can thus be used as surrogate processes for MSE computation via linear filtering techniques \cite{Anderson:1979}.

\subsubsection*{The MSE of the optimal linear estimator of $z_k$}

First, it is useful to consider the asymptotic prediction error of $q_k$.
In particular, we now compute the stationary solution $P$ of the Algebraic Riccati Equation (ARE), which gives the asymptotic prediction error variance of the Kalman filter applied to the state-space model describing $\tilde{z}_k$ and $\tilde{y}_k$ \cite{Anderson:1979}. The ARE reads:

\begin{eqnarray*}
P &=& a^2 P - \frac{a^2 P^2}{P + \sigma_n^2} + \sigma_\eta^2 \\
&=& a^2 P - \frac{a^2 P^2}{P + (\gamma^2 + \mu^2)(1 - \beta^2) + \sigma^2} + \beta^2 \ell^2.
\end{eqnarray*}

The closed-form solution to this equation is:
$$
P = \frac{Q - (1 - a^2)R + \sqrt{[(1 - a^2)R - Q]^2 + 4QR}}{2},
$$
where
$$
Q = \beta^2 \ell^2, \quad R = (\gamma^2 + \mu^2)(1 - \beta^2) + \sigma^2.
$$

This can be written more explicitly as:
{\small
\begin{equation}
\label{eqP}
\begin{aligned}
P = \frac{1}{2} \Big[ \, 
& \beta^2 \ell^2 
- (1 - a^2)\left( (\gamma^2 + \mu^2)(1 - \beta^2) + \sigma^2 \right) \\
& + \sqrt{
\left(
(1 - a^2)\left( (\gamma^2 + \mu^2)(1 - \beta^2) + \sigma^2 \right) - \beta^2 \ell^2
\right)^2
+ 4 \beta^2 \ell^2 \left( (\gamma^2 + \mu^2)(1 - \beta^2) + \sigma^2 \right)
}
\, \Big]
\end{aligned}
\end{equation}
}

This quantity $P$ represents the steady-state prediction error variance affecting $q_k$ asymptotically.

Consider now the optimal linear estimator of $z_k = (g_k+\mu)|w_k|$ based on all the
measurements $y_i = z_i + e_i$ collected up to instant $k$. Such estimator is unbiased and its MSE (variance)
coincides with that of the optimal linear estimator of $\tilde{z}_k = m + q_k + \varepsilon_k$ based on $\tilde{y}_i =
m + q_i + n_i, \ i=1,\ldots,k$, since these processes share the same first- and second-order moments. To compute such MSE, recall that, at steady state, before observing the measurement $\tilde{y}_k$, the prediction error
variance of $q_k$, which defines $\tilde{z}_k$ through \eqref{tildez}, was denoted by $P$. It follows that the variance of the prediction error of
$\tilde{z}_k$ is:
$$
P + \mathrm{Var}(\varepsilon_k) = P + (\gamma^2 + \mu^2)(1 - \beta^2).
$$

After receiving the new measurement
$$
\tilde{y}_k = m + q_k + \varepsilon_k + e_k = \tilde{z}_k + e_k,
$$
the asymptotic variance of the estimation error of $\tilde{z}_k$ is
$$
\mathrm{MSE}_2 = \left(\frac{1}{P + (\gamma^2 + \mu^2)(1 - \beta^2)} + \frac{1}{\sigma^2}\right)^{-1},
$$
with $P$ given in \eqref{eqP}. Thus, this is also the MSE affecting asymptotically
$z_k$ in terms of the original parameters:
$\mu$ (the asymptotic mean of $g_k$), $\ell^2$ (variance of the noise driving $g_k$),
$\beta$ (the mean of $|w_k|$),
$a$ (the AR(1) parameter regulating the correlation among the $g_k$),
$\sigma^2$ (output noise variance), and the steady-state variance of $g_k$:
$$
\gamma^2 := \frac{\ell^2}{1 - a^2}.
$$

\subsubsection*{MSE comparison: role of signs knowledge}

We are now in a position to understand how much improvement in intrinsic noise estimation can be obtained by exploiting knowledge of its signs.
According to the analysis performed in the previous sections,
we need to investigate the ratio:
$$
r :=\frac{\mathrm{MSE}_2}
{\mathrm{MSE}_1}
$$
where
$$
\mathrm{MSE}_1 =
\frac{(\mu^2 + \gamma^2)\sigma^2}
{\mu^2 + \gamma^2 + \sigma^2}.
$$
and
$$
\mathrm{MSE}_2 =
\frac{[P + (\mu^2 + \gamma^2)(1- \beta^2)]\sigma^2}
{P + (\mu^2 + \gamma^2)(1- \beta^2) + \sigma^2}
$$
with $P$ given in \eqref{eqP}. Small values of $r$ mean that signs knowledge has an important effect in reducing the estimation
error of the intrinsic noise. Recalling that $\beta:=\mathbb{E}[|w_k|]$, the following result then holds.

\begin{proposition}\label{Prop1}
One has
$$
r \in (1 - \beta^2, 1]
$$
and every value in this semi-open
interval can be achieved by appropriately selecting the system parameters.
\end{proposition}

Before proving this result, some comments are in order.  
In the case where $w_k$ is Gaussian, it holds that $\beta^2 = \frac{2}{\pi}$.
Therefore, the maximum possible MSE reduction obtained by using the second
estimator instead of the first one is considerable since 
$$
\inf r=1 - \frac{2}{\pi} \approx 0.37 \quad (\text{Gaussian case}),
$$
see also Fig. \ref{fig:MSE_comparison} which displays some profiles of $r$ as a function of the correlation parameter $a$.\\
If $w_k$ follows a Laplacian distribution, it holds that $\beta^2 = \frac{1}{2}$ so that
$$
\inf r=\frac{1}{2} \quad (\text{Laplacian case}).
$$
An additional example is a symmetric Bernoulli process where $w_k = \pm1$ with equal probability and one thus has $\beta^2 = 1$.
Therefore, the infimum of $r$ is zero in the
Bernoulli case, 
$$
\inf r=0 \quad (\text{Bernoulli case}),
$$
because $\mathrm{MSE}_2$ can be made arbitrarily small by increasing the correlation among the $g_k$.
Such improvements are possible because the second estimator (corresponding to the second stage of Trine) can exploit correlations among all output measurements, whereas the first one cannot.\\

{\bf{Proof of Proposition \ref{Prop1}}:}
We begin by proving that $r \le 1$, which is equivalent to:
\[
\frac{[P + (\mu^2 + \gamma^2)(1- \beta^2)]\sigma^2}
{P + (\mu^2 + \gamma^2)(1- \beta^2) + \sigma^2}
\le \frac{(\mu^2 + \gamma^2)\sigma^2}
{\mu^2 + \gamma^2 + \sigma^2}
.
\]
This simplifies to:
\[
P \le \beta^2(\mu^2 + \gamma^2) = :P_0.
\]
We can rewrite the ARE as:
\[
f(P ) := (1- a^2)(P - \beta^2 \gamma^2)(P - P_0 + \gamma^2 + \mu^2 + \sigma^2) + a^2 P^2 = 0.
\]
Since the coefficient of $P^2$ is 1 and the ARE has exactly one positive solution
(and one negative), we conclude that $P \le P_0$ if and only if $f(P_0) \ge 0$, which
yields:
\[
f(P_0) = \beta^2(1- a^2)\mu^2(\gamma^2 + \mu^2 + \sigma^2) + a^2 P_0^2 \ge 0,
\]
which is always true. The equality $P = P_0$ (and thus $r = 1$) occurs only when
$a = 0$ and $\mu = 0$ (i.e., $g_k$ is white noise and has zero mean).

We now prove that $r > 1 - \beta^2$ and that all values in the interval $(1 - \beta^2, 1)$ are attainable. 
Define:
\[
P_0 = \delta \sigma^2, \quad P = \epsilon \sigma^2, \quad \gamma^2 = \beta^2 \phi \sigma^2,
\]
which implies $\epsilon \le \delta$ from $P \le P_0$.
Then, the ARE becomes:
\[
a^2 \epsilon^2 + (1- a^2)
\left\{
[
(\beta^2 - 1)\delta + 1- \phi
]
\epsilon- \phi
[
(\beta^2 - 1)\delta - 1
]
\right\}
= 0
\]
which has to be solved in the unknown $\epsilon$.
Note that $\delta$ and $\phi$ depend only on $\mu^2, \gamma^2$, and $\sigma^2$, and can be selected independently of $a$. Although $\gamma^2 = \frac{\ell^2}{1 - a^2}$ seems to depend on $a$, for any fixed $\gamma^2$ we can always define $\ell^2(a, \gamma^2) := \gamma^2(1 - a^2)$. As $a \to 1^-$, it follows that $\epsilon \to 0^+$.

Now, the MSEs can be rewritten as:
\[
\mathrm{MSE}_1 =
\frac{P_0\sigma^{2}}
{P_0 + \beta^2 \sigma^2}
=
\frac{\delta}
{\delta + \beta^2}
\sigma^2,
\]
\[
\mathrm{MSE}_2 =
\frac{[P + P_0(
1/\beta^2 - 1)]\sigma^2}
{P + P_0(
1/\beta^2 - 1) + \sigma^2}
=
\frac{\epsilon+ ( 1/\beta^2 - 1)\delta}
{\epsilon+ ( 1/\beta^2 - 1)\delta + 1}
\sigma^2.
\]
Since $\mathrm{MSE}_2$ is an increasing function of $\epsilon$ and $\mathrm{MSE}_1$ does not depend on it,
the infimum of the ratio is attained as $\epsilon \to 0^+$. Define:
$$
\inf_{\epsilon \in (0,\delta]} r(\epsilon, \delta) := r(\delta) =\frac{\delta+\beta^2}{\delta+\frac{\beta^2}{1-\beta^2}}.
$$
This function is increasing in $\delta$, so its infimum is obtained as $\delta \to 0^+$, while
its supremum is obtained letting $\delta$ go to infinity,
yielding:
\begin{eqnarray*}
\inf_{\delta} r(\delta) &=& 1- \beta^2, \\
\sup_{\delta} r(\delta) &=&1.
\end{eqnarray*}
Therefore, for any $r \in (1 - \beta^2, 1)$, we can select $\delta > 0$ and let $\epsilon \to 0^+$ (i.e.,
$a \to 1^-$) to achieve it.
To confirm this, observe that:
\[
r(\epsilon, \delta) \ge 1- \beta^2
\]
for all $0 < \epsilon \le \delta$. Finally, recall also that $r=1$ is obtained with $a=\mu=0$ and this completes the proof.\\

In conclusion, the maximum performance improvement is achieved as $a \to 1^-$ and $\gamma^2, \mu^2 \ll \sigma^2$, whereas the worst-case occurs when $a = \mu=0$.

\begin{figure}[htbp]
    \centering
    \includegraphics[width=0.75\textwidth]{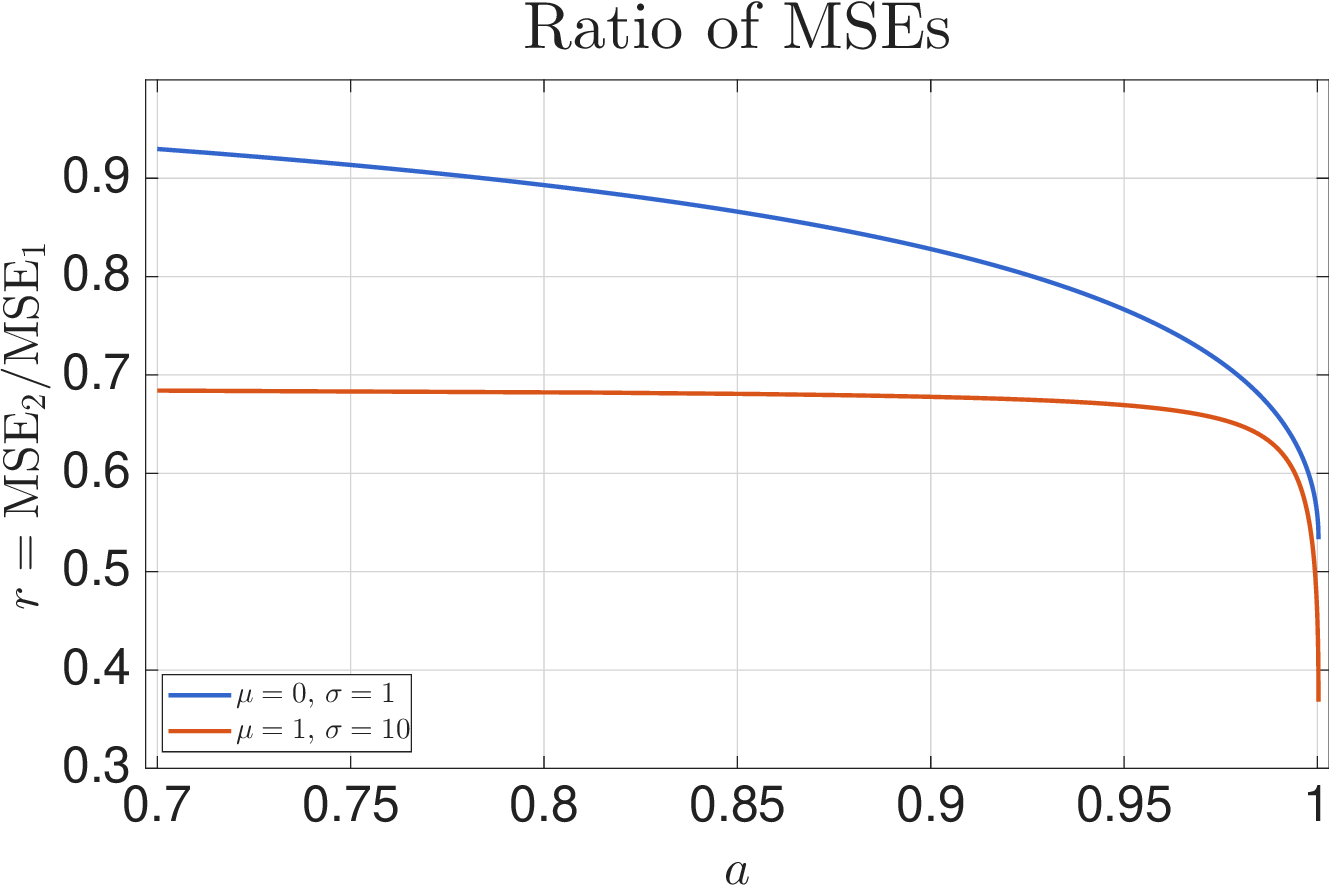}
    \caption{Comparison of the ratio $r = \frac{\mathrm{MSE}_2}{\mathrm{MSE}_1}$ as a function of the correlation parameter $a$ 
    for two parameter settings using Gaussian intrinsic noise. Smaller values correspond to larger performance gains for the estimator that leverages the signs of intrinsic noise realizations. Setting $\gamma=1,\beta^2 = \frac{2}{\pi}$ (Gaussian intrinsic noise case), the blue curve  then corresponds to $\mu=0,\sigma=1$ while the orange curve corresponds to 
    $\mu=1,\sigma=10$.}
    \label{fig:MSE_comparison}
\end{figure}

\subsubsection*{Interpretation of the result in the Trine setting}

In the Trine setting, the results obtained here support this conclusion. As shown in the derivation of the Trine algorithm in the first part of this Supplementary Material, and according to the Bayesian interpretation of regularization, if the postulated covariance for the SD profile $g$ is correct, the kernel obtained in the second stage possesses theoretical optimality. In fact, it enables the construction of the minimum-variance linear estimator of the intrinsic noise under the assumption of known signs. Therefore, if the sign estimates produced in the first stage are sufficiently accurate, the structured kernel can fully exploit the information encoded in the regularity of the variance profile, thereby improving the intrinsic noise estimates. The more regular the variance profile is, the lower the estimator’s MSE will be, as quantified in this section and illustrated in the examples in Fig. \ref{fig:MSE_comparison}.

\subsection*{Monte Carlo study}

We present an additional study concerning the Toggle switch example. 
This analysis further illustrates Trine’s robustness and the meaning of the theoretical findings outlined in the previous section.\\  
We perform a Monte Carlo study using the same Toggle switch model introduced in the main text, in which the system describes the evolution of two proteins, 
expressed from two mutually inhibiting promoters and modeled through stochastic differential equations \cite{Elston2001}.
In each run, we use  1000 data points and generate independent realizations of both the intrinsic noise and the output noise. The standard deviation of the output Gaussian noise is set as a varying fraction of the true state values, so that each run corresponds to a different coefficient of variation (CV\%).

We repeat the procedure for 10000 independent runs. For each run, we compute the norm of the realization of the output noise and compare it to the norm of the intrinsic noise realization. Based on the ratio of these norms, we categorize the runs into different bins.
Within each bin, we report the average percentage fit, over all runs that fall into the corresponding bin,
between the estimated $\widehat{G}$ reported in \eqref{widehatG} and true profile $G$ of the intrinsic noise standard deviation. Recall that the fit is defined by
$$
100\Big( 1 - \frac{\|G-\widehat{G}\|}{\|G\|} \Big),
$$
so that higher fit percentage indicates better estimator performance.

In addition to the full \textbf{Trine} algorithm, we compare two alternative versions:
\begin{itemize}
    \item \textbf{Trine$^{\mathbf{u}}$}: a version in which the third stage (estimation of the intrinsic noise variance profile) uses directly the intrinsic noise estimates in \eqref{EstINFirstStage} coming from the first stage, i.e., obtained without the structured kernel (the superscript $u$ indeed stands for \emph{unstructured}).
    \item \textbf{Oracle}: a baseline version in which the third stage uses the true intrinsic noise realizations. This serves as a benchmark to evaluate the performance ceiling of the estimator.
\end{itemize}

We summarize the performance of the estimators in Table~\ref{TabMC}. Each row corresponds to a bin defined by the ratio between the norm of  the realizations of the output and intrinsic noise. Within each bin, we report the average percentage fit for Trine, Trine$^{\mathbf{u}}$ and Oracle. 
Each average is computed over at least 1000 Monte Carlo runs.

\begin{table}[h!]
\centering
\caption{\textbf{Monte Carlo results across bins of output-to-intrinsic noise ratio.} Each bin corresponds to a range of values for the ratio $\|e\|/\|n\|$, where $e$ is the output noise realization and $n$ is the intrinsic noise realization. In each bin, the reported values are the average percentage fit between the estimated and true intrinsic noise standard deviation profile. A higher fit indicates better performance.}
\vspace{0.5em}
\begin{minipage}{0.7\textwidth}\label{TabMC}
\scalebox{1}{
\renewcommand{\arraystretch}{1.6}
\begin{tabular}{|c|c|c|c|c|}
\hline
\textbf{Bin range} ($\|e\|/\|n\|$) & \textbf{Trine} & \textbf{Trine$^{\mathbf{u}}$} & \textbf{Oracle} \\
\hline 
$[0.0,\ 0.1)$   & \textbf{91.4} & 73.5 & 92.2 \\ 
$[0.1,\ 0.2)$   & \textbf{90.4} & 71.6 & 92.1 \\ 
$[0.2,\ 0.3)$   & \textbf{87.4} & 70.8 & 92.2 \\ 
$[0.3,\ 0.4)$   & \textbf{82.1} & 67.9 & 92.2 \\ 
$[0.4,\ 0.5)$   & \textbf{74.8} & 67.3 & 92.2 \\ 
$[0.5,\ 0.6)$   & \textbf{65.2} & 68.3 & 92.1 \\ 
$[0.6,\ 1.0]$   & \textbf{42.2} & 61.5 & 91.8 \\ 
\hline
\end{tabular}
}
\end{minipage}
\end{table}

It is remarkable that when the output noise does not dominate over the intrinsic noise, \textbf{Trine} achieves a performance very close to that of the oracle. Moreover, it largely outperforms \textbf{Trine$^{u}$} because the intrinsic noise signs are accurately estimated, so that the theoretical results presented in the previous section suggest the structured kernel in the second stage can be effectively exploited.

Note, however, that when the output noise increases and effectively masks the intrinsic noise, it may be preferable to use \textbf{Trine$^{u}$}. In this regime, the sign estimates become too imprecise, and the theoretical properties of the structured estimator discussed earlier can no longer be exploited. In fact, the structured kernel becomes too complex to be estimated reliably, leading to a degradation in Trine’s performance.
To this regard, it is interesting to note that the first step of Trine can also be used to estimate the relative strength between output and intrinsic noise. Indeed, the first phase returns the parameter $\rho_n$, which can be interpreted as an estimate of the average variance of the intrinsic noise. This can be compared with the average $\rho_e$ of the diagonal elements of $\Sigma_e$, which represents the mean variance of the output noise. A large ratio $\sqrt{\rho_e}/\sqrt{\rho_n}$ suggests that the results from \textbf{Trine$^{u}$} should be considered.




\subsection*{Toggle Switch Simulation Details}

The two-dimensional state vector contains the two proteins $x_1$ and $x_2$ coming from two different promoters mutually inhibiting each other. Their dynamics are regulated by the following system of stochastic differential equations~\cite{Elston2001}:
\[
\dot{x}(t) = f(x(t))\,dt + g(x(t))\,w(t),
\]
where the drift $f(x)$ contains the two deterministic functions
\begin{footnotesize}
\[
\begin{aligned}
f_1(x_1,x_2)
&= \frac{b + x_1^2}{\,b + x_1^2 + x_2^2\,} - x_1
\\
&-\frac{1}{\kappa}\,
\frac{2\,x_1 x_2 (x_1 + x_2)\,\Big[
(x_1 - 1)(b + x_1^2)(2b + x_1^2)
+ x_1 x_2^2\big(3b + x_1(2x_1 - 1)\big)
+ x_1 x_2^4
\Big]}
{\,b\,\big(b + x_1^2 + x_2^2\big)^4},
\\[6pt]
f_2(x_1,x_2)
&= \frac{b + x_2^2}{\,b + x_1^2 + x_2^2\,} - x_2
\\
&- \frac{1}{\kappa}\,
\frac{2\,x_1 x_2 (x_1 + x_2)\,\Big[
(x_2 - 1)(b + x_2^2)(2b + x_2^2)
+ x_2 x_1^2\big(3b + x_2(2x_2 - 1)\big)
+ x_2 x_1^4
\Big]}
{\,b\,\big(b + x_1^2 + x_2^2\big)^4}.
\end{aligned}
\]
\end{footnotesize}

The components $w_1,w_2$ of $w$ 
are independent white Gaussian noises with unit variance. 
Finally, the state-dependent covariance of the intrinsic noise $g(x(t))w(t)$ is

\begin{small}
\[
Q(x_1,x_2) := g(x)\,g(x)^\top=
\begin{bmatrix}
Q_1(x_1,x_2) & Q_{12}(x_1,x_2)\\[2pt]
Q_{12}(x_1,x_2) & Q_2(x_1,x_2)
\end{bmatrix},
\]
with entries
\[
\begin{aligned}
Q_1(x_1,x_2)
&= \frac{1}{m_0}\left(
\frac{b + x_1^2}{\,b + x_1^2 + x_2^2\,} + x_1
\right)
+ \frac{1}{\kappa}\,
\frac{x_2^2\left(b^2 + 2b\,x_1^2 + x_1^2 x_2^2 + x_1^4\right)}
{\,b\,\big(b + x_1^2 + x_2^2\big)^3},
\\[8pt]
Q_2(x_1,x_2)
&= \frac{1}{m_0}\left(
\frac{b + x_2^2}{\,b + x_1^2 + x_2^2\,} + x_2
\right)
+ \frac{1}{\kappa}\,
\frac{x_1^2\left(b^2 + 2b\,x_2^2 + x_1^2 x_2^2 + x_2^4\right)}
{\,b\,\big(b + x_1^2 + x_2^2\big)^3},
\\[8pt]
Q_{12}(x_1,x_2)
&= \frac{1}{\kappa}\,
\frac{x_1^2 x_2^2 \left(2b + x_1^2 + x_2^2\right)}
{\,b\,\big(b + x_1^2 + x_2^2\big)^3}.
\end{aligned}
\]
\end{small}

Finally, model parameters used to simulate the data
are reported in Table \ref{TabTS}.

\begin{table}[h!]
\centering
\caption{Toggle switch --- model parameters}
\begin{tabular}{lcl}\label{TabTS}
Parameter & Value & Description \\
\hline
$\theta$     & $1\times 10^{4}\ \mathrm{molec/s}$ & Active production rate \\
$\alpha$     & $1000\ \mathrm{s^{-1}}$           & Degradation rate \\
$\beta$      & $50$                              & Cooperativity \\
$\delta$     & $0.75\ \mathrm{s^{-1}}$           & Switch ON rate \\
$\alpha_1$   & $\alpha$                & Degradation rate [min$^{-1}$] \\
$\delta_1$   & $\delta$                & Switch ON rate [min$^{-1}$] \\
$K$          & $0.01$                            & Parameter $K$ \\
$b$          & $\displaystyle \frac{\beta\,\theta\,\delta^{2}}{\alpha_1^{2}}$ & Dimensionless parameter $b$ \\
$\kappa$     & $\displaystyle \frac{K\,\alpha_1^{2}}{\theta\,\delta^{3}}$     & Dimensionless parameter $\kappa$ \\
$m_0$        & $1000$                            & Molecular scale \\
\hline
\end{tabular}
\end{table}

\end{document}